\documentclass{article}

\usepackage[preprint]{neurips_2026}

\usepackage[utf8]{inputenc} %
\usepackage[T1]{fontenc}    %
\usepackage{hyperref}       %
\usepackage{url}            %
\usepackage{booktabs}       %
\usepackage{amsmath}        %
\usepackage{amsfonts}       %
\usepackage{nicefrac}       %
\usepackage{microtype}      %
\usepackage[table]{xcolor}  %
\usepackage{enumitem}       %
\usepackage{tikz}           %
\usepackage{makecell}       %
\usepackage{multirow}       %
\usepackage{array}          %
\usepackage{threeparttable} %
\usepackage{float}          %
\usepackage{pifont}         %
\usepackage{pdflscape}      %
\usepackage[skins,breakable,listings]{tcolorbox} %
\usepackage{listings}                            %
\usepackage{listingsutf8}                        %

\newtcolorbox{promptbox}[1]{%
  enhanced, breakable,
  colback=black!4, colframe=black!18,
  boxrule=0.4pt, arc=2pt,
  left=6pt, right=6pt, top=4pt, bottom=4pt,
  fontupper=\footnotesize\ttfamily,
  before skip=6pt, after skip=6pt,
  attach boxed title to bottom right={yshift=-1ex, xshift=-6pt},
  boxed title style={colback=black!10, colframe=black!18, sharp corners, boxrule=0pt},
  coltitle=black!60,
  fonttitle=\scriptsize\ttfamily,
  title=#1,
}

\newtcblisting{promptlisting}[1]{%
  enhanced, breakable,
  colback=black!4, colframe=black!18,
  boxrule=0.4pt, arc=2pt,
  left=6pt, right=6pt, top=4pt, bottom=4pt,
  before skip=6pt, after skip=6pt,
  attach boxed title to bottom right={yshift=-1ex, xshift=-6pt},
  boxed title style={colback=black!10, colframe=black!18, sharp corners, boxrule=0pt},
  coltitle=black!60,
  fonttitle=\scriptsize\ttfamily,
  title={\detokenize{#1}},
  listing only,
  listing options={
    basicstyle=\scriptsize\ttfamily,
    breaklines=true,
    breakatwhitespace=true,
    columns=fullflexible,
    showstringspaces=false,
    aboveskip=0pt, belowskip=0pt,
    inputencoding=utf8/latin1,
    extendedchars=true,
    literate={—}{{---}}1 {–}{{--}}1 {“}{{``}}1 {”}{{''}}1 {‘}{{`}}1 {’}{{'}}1,
  },
}

\newcommand{\promptfile}[1]{%
  \tcbinputlisting{
    enhanced, breakable,
    colback=black!4, colframe=black!18,
    boxrule=0.4pt, arc=2pt,
    left=6pt, right=6pt, top=4pt, bottom=4pt,
    before skip=6pt, after skip=6pt,
    attach boxed title to bottom right={yshift=-1ex, xshift=-6pt},
    boxed title style={colback=black!10, colframe=black!18, sharp corners, boxrule=0pt},
    coltitle=black!60,
    fonttitle=\scriptsize\ttfamily,
    title={\detokenize{#1}},
    listing only,
    listing file={#1},
    listing options={
      basicstyle=\scriptsize\ttfamily,
      breaklines=true,
      breakatwhitespace=true,
      columns=fullflexible,
      showstringspaces=false,
      aboveskip=0pt, belowskip=0pt,
      inputencoding=utf8/latin1,
      extendedchars=true,
      literate={—}{{---}}1 {–}{{--}}1 {“}{{``}}1 {”}{{''}}1 {‘}{{`}}1 {’}{{'}}1,
    },
  }%
}
\usetikzlibrary{positioning, shapes.geometric, arrows.meta}
\setlist{itemsep=1pt, topsep=2pt, parsep=0pt}

\usepackage{comment}
\IfFileExists{dsfont.sty}{\usepackage{dsfont}}{}
\providecommand{\mathds}[1]{\mathbf{#1}}

\usepackage{titlesec}   %
\usepackage{caption}    %
\usepackage{etoolbox}   %

\titlespacing*{\section}      {0pt}{1.2ex plus 0.3ex minus 0.2ex}{0.6ex plus 0.2ex}
\titlespacing*{\subsection}   {0pt}{1.0ex plus 0.3ex minus 0.2ex}{0.4ex plus 0.2ex}
\titlespacing*{\subsubsection}{0pt}{0.8ex plus 0.2ex minus 0.2ex}{0.3ex plus 0.1ex}
\titlespacing*{\paragraph}    {0pt}{0.6ex plus 0.2ex minus 0.1ex}{0.8em}

\usepackage{enumitem}
\setlist{topsep=2pt, partopsep=0pt, itemsep=1pt, parsep=1pt, leftmargin=*}

\title{CompoSET: A Controlled Single-Edit Benchmark for Vision-Language Compositionality}

\author{%
  Laure Delisle \\
  California Institute of Technology
  \And
  Pietro Perona \\
  California Institute of Technology
}

\begin{document}

\raggedbottom

\maketitle

\begin{abstract}
\hyphenpenalty=10000
\exhyphenpenalty=10000
Vision-language compositionality is the ability to identify which properties, relations, quantities, and states bind to which entities in both an image and a caption, and to match those bindings across modalities. Existing benchmarks evaluate this ability through image-text matching:
a model ranks candidate image-caption match similarities and succeeds when the correct match comes on top. In such datasets, paired images differ by more than a single binding: scene, layout, and content all differ, with 97\% of pixels changing on average. Such pairs do not isolate compositionality: results are confounded by global image differences. We introduce Compositional Single-EdiT (CompoSET), a benchmark dataset in which each base image-caption pair has an edited counterpart, or foil, that preserves the scene but differs by a single controlled edit, such as color, count, spatial relation, or object state. Base and foil images preserve naturalistic scene complexity yet are visually near-identical: only 8\% of pixels change on average.  Our dataset has many foils per base scene ($\sim$22) in order to separate scene difficulty from edit difficulty, a distinction one-foil-per-base benchmarks cannot make. CompoSET contains 1,776 instances across 80 naturalistic scenes, spanning 16 edit types in 4 semantic categories, with captions at 3 verbosity levels. We test 22 dual-encoder VLMs and find that all models perform at or below chance level on CompoSET, despite scoring well on prior benchmarks. This suggests that apparent progress under prior image-text matching protocols may be an artifact arising from confounds in benchmark design, and that compositional understanding in dual-encoder VLMs is far from solved. CompoSET addresses the confounds and offers a better controlled benchmark to measure future progress.

\medskip\noindent
{\small
\raisebox{-0.2ex}{\includegraphics[height=1em]{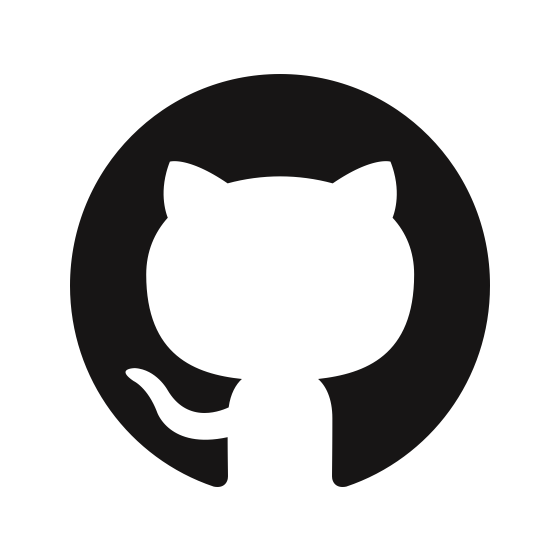}}\hspace{0.3em}\textbf{Code:} \url{https://github.com/laure-delisle/CompoSET}\\
\raisebox{-0.2ex}{\includegraphics[height=1em]{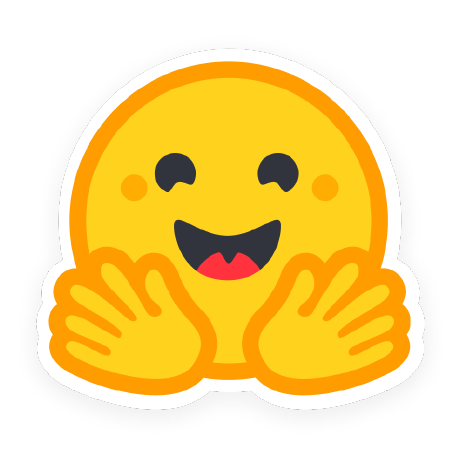}}\hspace{0.3em}\textbf{Dataset:} \url{https://huggingface.co/datasets/CompoSET/CompoSET}
}

\end{abstract}

\section{Introduction}
\label{sec:intro}

Compositional understanding in vision-language models is often tested through image-text matching: can a model match captions to images based on how attributes, relations, quantities, states, and roles bind to objects? A model that cannot distinguish ``a red tin next to a blue bottle'' from ``a blue tin next to a red bottle'' has learned co-occurrence without binding. We study this capability in dual-encoder VLMs, which remain central to scalable retrieval: images and captions are embedded once, indexed, and searched by similarity. Generative VLMs can answer richer per-query questions, but their scoring is conditional and expensive; they do not replace pre-indexed retrieval embeddings. If the embedding does not preserve the distinction between two minimally different scenes, retrieval is unlikely to rank them correctly.

Existing compositionality benchmarks aim to test this capability, but their image-text matching evaluations often rely on structurally dissimilar image pairs. In prior benchmarks, paired images differ in scene, layout, and content, with 97\% of pixels changing on average. This is a measurement problem: matching scores can reflect global image differences rather than sensitivity to the targeted composition.

\begin{figure}[t]
\centering
\includegraphics[width=\textwidth]{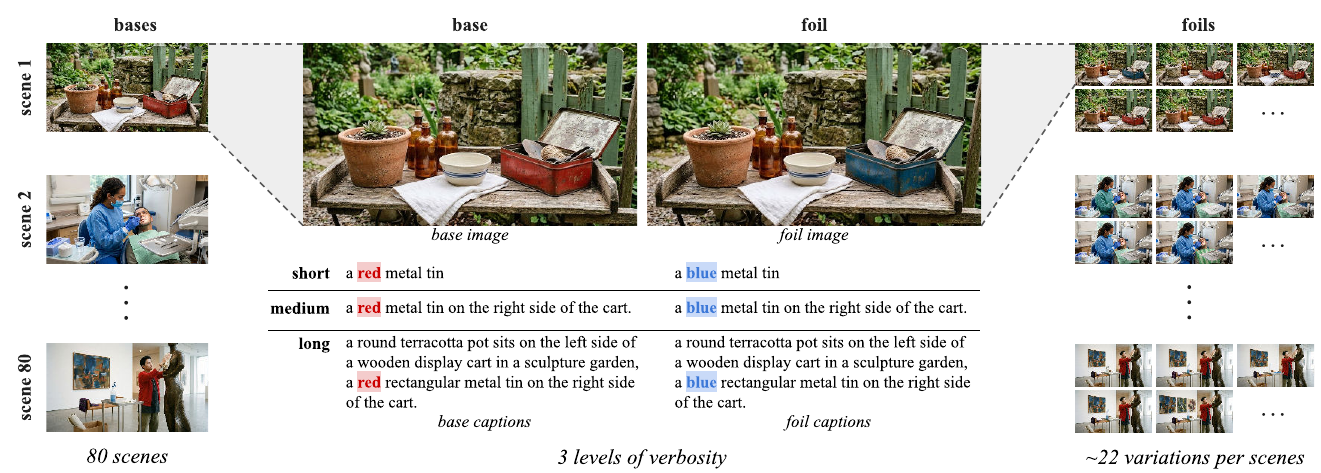}
\caption{\textbf{Compositionality experiments require comparing images that differ by one named factor.} CompoSET contains 80 naturalistic multi-object {\em base} images {\bf (Left)} that are varied to obtain 1,776 instances, the {\em foils} {\bf (Right)} (22 foils per base). Each base-foil pair differs by exactly one compositional edit which is labeled in the captions {\bf (Middle)} (Sec.\ref{sec:benchmark}). Captions come at three levels of verbosity to test sensitivity to context. In the experiments models compute the similarity between each one of two related images (a base image and one of its foils) and the corresponding two captions, and vice-versa, for a total of four comparisons (Sec.\ref{sec:experiments}). }
\label{fig:teaser}
\end{figure}

A controlled benchmark should satisfy seven design criteria:
\begin{enumerate}[leftmargin=*, itemsep=1pt, topsep=2pt]
    \item \textbf{Naturalistic scenes}, so conclusions transfer beyond toy renderings.
    \item \textbf{Multi-object scenes}, so binding attributes, relations, quantities, states, and roles to the correct entities is nontrivial.
    \item \textbf{Same-scene single-edit pairs}, so the compositional variable is isolated.
    \item \textbf{Multiple foils per base scene}, so scene difficulty can be separated from edit difficulty.
    \item \textbf{A meaningful edit taxonomy}, so failures can be localized by semantic type.
    \item \textbf{Caption verbosity levels}, to test whether retrieval remains sensitive to the compositional edit as surrounding context dilutes the edit phrase.
    \item \textbf{Bidirectional retrieval scoring}, so both image-to-text and text-to-image discrimination are measured.
\end{enumerate}

Prior benchmarks satisfy important subsets of these criteria, but not the full set. Text-only foils cannot test image-side compositional discrimination; different-scene image foils confound composition with scene-level variation; and existing same-scene benchmarks are limited in scene complexity, edit coverage, or foils per base. Table~\ref{tab:bench-comparison} summarizes the gap: CompoSET is the only benchmark we evaluate that satisfies all seven criteria.

We introduce Compositional Single-EdiT (CompoSET), a benchmark dataset built to satisfy this full controlled-evaluation specification. Each instance pairs a base image-caption pair with an edited counterpart, or foil, that preserves the scene while changing a single visual element and its corresponding caption. CompoSET contains 1,776 instances across 80 naturalistic scenes, with $\sim$22 foils per base, 16 edit types, and 3 caption verbosity levels. Together, these choices hold the scene fixed, vary one semantic factor, and make failures measurable edit by edit and scene by scene.

Across 22 dual-encoder VLMs, all models perform at or below chance level on CompoSET, despite high scores on prior benchmarks. Models that appear compositionally competent under correlational comparisons fail under controlled evaluation, showing that apparent progress in dual-encoder compositionality is largely an artifact of benchmark design.

In summary, we make three contributions: we frame compositionality evaluation as controlled single-edit experimentation rather than correlational comparison; we introduce CompoSET, a benchmark dataset satisfying all seven criteria above; and we show that 22 dual-encoder VLMs perform at or below chance under this controlled protocol.

\begin{table}[t]
\centering
\caption{\textbf{CompoSET satisfies all criteria; no prior benchmark does.} The ten compositionality benchmarks we evaluate, scored on six of the seven criteria from Sec.\ref{sec:intro}. The remaining criterion, multi-object scene complexity (2), is quantified in Sec.\ref{sec:stats-objcount} and Table~\ref{tab:objcount-distrib}. \textcolor[HTML]{4A90D9}{\checkmark}\ marks a criterion satisfied, \ding{53}\ unsatisfied, ``--'' not applicable (text-only foils). \emph{Naturalistic}: photo-like real-world imagery. \emph{Same-scene}: base and foil depict the same visual scene. \emph{Edit types}: number of distinct compositional axes probed. \emph{Verbosity}: caption verbosity levels per instance. Best-on-criterion cells in \textbf{bold}.}
\label{tab:bench-comparison}
\small
\setlength{\tabcolsep}{4pt}
\makebox[\textwidth][c]{%
\begin{tabular}{lcccccc}
\toprule
Benchmark & Naturalistic & Same-scene & Foils/base & Edit types & Verbosity & Bidir.\\
\midrule
\rowcolor{black!7}
\textbf{CompoSET (ours)} & \textbf{\textcolor[HTML]{4A90D9}{\Large\textbf{\checkmark}}} & \textbf{\textcolor[HTML]{4A90D9}{\Large\textbf{\checkmark}}} & \textbf{$\sim$22} & \textbf{16} & \textbf{3} & \textbf{\textcolor[HTML]{4A90D9}{\Large\textbf{\checkmark}}}\\
Winoground~\cite{thrush2022winoground} & \textcolor[HTML]{4A90D9}{\Large\textbf{\checkmark}} & \ding{53} & 1 & 3 & 1 & \textcolor[HTML]{4A90D9}{\Large\textbf{\checkmark}}\\
BiVLC~\cite{miranda2024bivlc} & \ding{53} & \ding{53} & 1 & 5 & 1 & \textcolor[HTML]{4A90D9}{\Large\textbf{\checkmark}}\\
SugarCrepe~\cite{hsieh2023sugarcrepe} & \textcolor[HTML]{4A90D9}{\Large\textbf{\checkmark}} & -- & 1 & 7 & 1 & --\\
SugarCrepe++~\cite{dumpala2024sugarcrepepp} & \textcolor[HTML]{4A90D9}{\Large\textbf{\checkmark}} & -- & 1 & 5 & 1 & --\\
What's Up~\cite{kamath2023whatsup} & \textcolor[HTML]{4A90D9}{\Large\textbf{\checkmark}} & \textcolor[HTML]{4A90D9}{\Large\textbf{\checkmark}} & 3 & 4 & 1 & --\\
ColorBench-robust~\cite{liang2025colorbench} & \textcolor[HTML]{4A90D9}{\Large\textbf{\checkmark}} & \textcolor[HTML]{4A90D9}{\Large\textbf{\checkmark}} & 9 & 1 & 1 & --\\
NaturalBench~\cite{li2024naturalbench} & \textcolor[HTML]{4A90D9}{\Large\textbf{\checkmark}} & \ding{53} & 1 & -- & 1 & --\\
MMVP~\cite{tong2024mmvp} & \textcolor[HTML]{4A90D9}{\Large\textbf{\checkmark}} & \ding{53} & 1 & 9 & 1 & \textcolor[HTML]{4A90D9}{\Large\textbf{\checkmark}}\\
ColorSwap~\cite{burapacheep2024colorswap} & \ding{53} & \ding{53} & 1 & 1 & 1 & \textcolor[HTML]{4A90D9}{\Large\textbf{\checkmark}}\\
\bottomrule
\end{tabular}%
}
\end{table}

\section{Related Work}
\label{sec:related}

Compositionality benchmarks for VLMs face a central design tension: they need alternatives that differ in composition, but not in everything else. Winoground~\cite{thrush2022winoground} made this tension visible with a $2\times2$ task: two captions contain the same words in different compositions, and two images must be matched to them. The captions are close to ideal compositional foils. The images, however, were collected by linguistic experts searching Getty Images for pictures that matched each caption separately. As a result, the two images need not share a scene, layout, style, or visual content beyond the semantics needed to satisfy their respective captions. Winoground therefore established the right evaluation question, but not a controlled image-side experiment: success and failure can still be entangled with global image differences. Later analysis reinforced this point by showing that Winoground difficulty is heterogeneous, mixing compositional matching with commonsense, visual ambiguity, and fine-grained recognition~\cite{diwan2022winoground}.

\textbf{Caption-perturbation benchmarks.}
Caption-perturbation benchmarks address scale and systematic linguistic control. ARO~\cite{yuksekgonul2023bow} and CREPE~\cite{ma2023crepe} construct hard negative captions to test relation, attribution, word order, systematicity, and productivity at scale; SugarCrepe~\cite{hsieh2023sugarcrepe} and SugarCrepe++~\cite{dumpala2024sugarcrepepp} improve negative-caption quality by reducing language-only artifacts through more fluent perturbations, adversarial refinement, and human filtering. VALSE~\cite{parcalabescu2022valse} and BLA~\cite{chen2023bla} similarly ask whether a model prefers a correct caption over a linguistically perturbed alternative. These benchmarks take a different cut at the same problem: hold the image fixed, perturb the caption, and generate many controlled linguistic contrasts. But the corresponding visual alternative is never constructed: the image never changes, so these benchmarks cannot test whether the image representation preserves that alternative.

\textbf{Image-side alternatives without same-scene control.}
A third line adds image-side alternatives at larger scale, but often without same-scene control. BiVLC~\cite{miranda2024bivlc} extends caption-perturbation benchmarks by generating negative images from negative captions, bringing scalable text perturbations into a bidirectional $2\times2$ retrieval format. ColorSwap~\cite{burapacheep2024colorswap}, NaturalBench~\cite{li2024naturalbench}, MMVP~\cite{tong2024mmvp}, and Cola~\cite{ray2023cola} also introduce image-side contrasts. These benchmarks broaden evaluation beyond caption discrimination, but the alternative image is usually a separately sourced or generated image rather than a controlled edit of the same scene. It may differ in scene, layout, content, style, or generation artifacts, so matching scores can reflect global visual differences rather than the intended compositional contrast.

\textbf{Same-scene control.}
The closest prior work moves toward same-scene control, but remains narrow in coverage. What's Up~\cite{kamath2023whatsup} tests spatial relations in two-object scenes; ColorBench-robust~\cite{liang2025colorbench} tests color only; SPEC~\cite{peng2024spec} and EqBen~\cite{wang2023eqben} include controlled synthetic subsets, trading away naturalistic multi-object complexity; CounterCurate~\cite{zhang2024countercurate} provides controlled axes but only one foil per base; and DreamSim/NIGHTS~\cite{fu2023dreamsim} is closest in image construction but evaluates perceptual similarity rather than image-text compositional matching. Image-editing datasets such as InstructPix2Pix~\cite{brooks2023instructpix2pix} provide useful edit supervision, but target instruction following or edit fidelity rather than bidirectional compositional image-text matching with controlled foils. These benchmarks show that controlled same-scene foils are possible, but none satisfy all seven criteria from Section~\ref{sec:intro} (Table~\ref{tab:bench-comparison}).

\textbf{Limits of CLIP-like embeddings.}
CompoSET connects to work on the limits of CLIP-like embeddings. MMVP~\cite{tong2024mmvp} identifies CLIP-blind image pairs, visual differences obvious to humans but compressed in CLIP embedding space, and shows that these failures transfer to CLIP-based multimodal models. Recent theoretical work argues that a single global cosine embedding space cannot simultaneously preserve content, attribute binding, spatial relations, and negation under CLIP-like assumptions~\cite{kang2025clipideal}. CompoSET brings this question into controlled evaluation: it does not identify the internal mechanism behind a failure, but measures whether current dual-encoder embeddings preserve single-edit distinctions in naturalistic image-text matching.

\textbf{Instruction-based image editing.}
Instruction-based Image Editing (IIE) models modify a source image according to a text instruction (e.g., ``make the landscape snowy''). Edits either target a single object (e.g., add, remove, count, color, material) or transform the entire image (e.g., style transfer, background replacement, denoising). Training sets contain triplets (source image, instruction, edited image)---in most datasets, target images are synthetically generated~\cite{zhang2023magicbrush,brooks2023instructpix2pix,zhao2024ultraedit,yu2025anyedit,ye2025imgedit}. Datasets with manual edits are limited to tone/removal tasks~\cite{shi2020gier} or single subjects~\cite{sushko2025realedit}. In contrast, many IIE benchmarks provide no target image~\cite{sheynin2024emuedit,ma2024i2ebench,ye2025imgedit,liu2025step1xedit}, since a single ground truth penalizes valid alternative edits. Instead, they evaluate instruction adherence and source preservation using pixel metrics, embedding similarities, or LLM judges.
Ultimately, IIE and compositionality benchmarks serve different purposes: they evaluate different models (generators vs.\ encoders) on distinct data (edit instructions vs.\ captions). Repurposing IIE data for compositionality is nontrivial, as it requires ground-truth targets (which many benchmarks lack), captioning both images, and strict controls to ensure that images and captions differ solely by the compositional edit.

\textbf{Composed image retrieval.}
Composed Image Retrieval (CIR) models retrieve a target image from a candidate set using a multimodal query: a reference image paired with a modifying text (e.g., ``is shorter and has no sleeves'')~\cite{song2026cirsurvey}. Retrieval is measured by Recall@$K$. Some benchmarks add hard negatives~\cite{mahadev2026pinpoint}, including ones edited from the reference image~\cite{song2026edir}. In CIRR, CIRCO, and Fashion IQ~\cite{liu2021cirr,baldrati2023circo,wu2021fashioniq}, reference and target are photographed separately, so they differ in unconstrained ways beyond the text. Editing-based datasets instead build targets by editing the reference, but add unrequested edits to increase retrieval difficulty~\cite{song2026edir}. CIR targets are therefore ill-suited for compositionality benchmarks. Adapting CIR data would require regenerating targets so they differ from the reference only by the compositional edit, and captioning both images.

\textbf{From editing data to a bidirectional test.}
CompoSET poses a third task, compositional retrieval: does a frozen embedding preserve a single controlled distinction, from image to text and from text to image? Editing and CIR data could seed such a test in three steps. First, keep only same-scene pairs with an object-level edit; this excludes most CIR pairs and all whole-image edits. Second, write a caption for each image that differs from its partner only in the edited phrase. A caption for the edited image alone supports only a one-sided text-to-image test; both captions are needed for image-to-text and for the joint score. Third, verify every pair for unintended changes, since generated edits drift. CompoSET performs these three steps on every instance.

\textbf{Positioning.}
CompoSET occupies the gap between lexical control and visual control. It keeps the $2\times2$ matching structure, but replaces separately sourced image alternatives with same-scene single edits. It keeps naturalistic multi-object scenes, but changes only one semantic factor. It gives each scene many foils, so failures can be analyzed by both scene and edit. This combination lets CompoSET separate controlled compositional measurement from apparent progress under global image confounds.

\section{The CompoSET Benchmark}
\label{sec:benchmark}

CompoSET is a benchmark for compositionality in vision-language models. We evaluate it through dual-encoder image-text matching. Each instance contains a base image with its caption and a foil image with its caption. The foil preserves the base scene while changing one semantic element ---\,an attribute, object, relation, quantity, or state\,--- together with the corresponding caption phrase. Images are synthetic but natural-looking, generated with frontier vision-language and image-generation models; every instance is human-reviewed (§\ref{sec:qc}).

\paragraph{Benchmark structure.} Each instance contains two image-caption pairs: a base pair and a foil pair. The evaluation asks whether a model assigns higher similarity to the matching pairs than to the mismatched alternatives, making the single edit the variable under test. CompoSET contains 1{,}776 instances from 80 naturalistic base scenes; each scene contributes at least 15 foils, with 22.2 foils on average.

\paragraph{Edit taxonomy.}\label{sec:taxonomy} Foils are drawn from a taxonomy of 16 edit types grouped into four semantic categories: Identity \& Appearance, Quantity, Spatial Relations, and Activity \& State. Table~\ref{tab:taxonomy} lists the full taxonomy and per-type counts. The taxonomy is organized by both semantic category and operation. A \emph{modify} edit changes one property of one entity, for example changing a metal tin from red to blue. A \emph{swap} edit exchanges a property or role between two entities, so the same ingredients appear in both base and foil but their assignment changes. Modify edits test sensitivity to a single changed assignment; swap edits remove simple presence cues and require resolving which entity carries which property or role. We exclude visually subjective or scene-level edits that cannot be controlled cleanly: age, gender, emotion, size, aesthetics, lighting, weather, and location. The structurally rare \texttt{swap\_role} edit requires at least two agents in the scene and is therefore less frequent than other edit types.

\begin{table}[t]
\centering
\caption{\textbf{CompoSET's 16 edit types span four semantic categories; prior benchmarks probe at most nine.} Per-type instance counts ($n$) in parentheses.}
\label{tab:taxonomy}
\footnotesize
\setlength{\tabcolsep}{4pt}
\begin{tabular}{@{}>{\raggedright\arraybackslash}p{0.24\linewidth}>{\raggedright\arraybackslash}p{0.72\linewidth}@{}}
\toprule
\textbf{Category} & \textbf{Edit types} ($n$) \\
\midrule
Identity \& Appearance (\emph{what}) &
\texttt{modify\_color} (206), \texttt{modify\_pattern} (101), \texttt{modify\_material} (161), \texttt{modify\_shape} (103), \texttt{modify\_transparency} (100), \texttt{modify\_object} (101), \texttt{swap\_color} (104), \texttt{swap\_material} (101) \\
\addlinespace
Quantity (\emph{how many}) &
\texttt{modify\_count} (100), \texttt{modify\_presence} (104) \\
\addlinespace
Spatial Relations (\emph{where}) &
\texttt{modify\_spatial} (99), \texttt{swap\_spatial} (100) \\
\addlinespace
Activity \& State (\emph{how}) &
\texttt{modify\_action} (101), \texttt{modify\_pose} (104), \texttt{modify\_state} (146), \texttt{swap\_role} (45) \\
\bottomrule
\end{tabular}
\end{table}

\paragraph{Caption verbosity.}\label{sec:verbosity}\label{sec:density} Each instance includes paired captions at three verbosity levels: \emph{short}, \emph{medium}, and \emph{long}. Short captions name the edited entity and property, for example ``a [red/blue] metal tin.'' Medium captions add a local anchor, for example ``a [red/blue] metal tin on the right side of the cart.'' Long captions add scene context while preserving the same edit phrase, for example ``A round terracotta pot sits on the left side of a wooden display cart in a sculpture garden, with a [red/blue] rectangular metal tin on the right side of the cart.'' The edit phrase stays fixed across verbosity levels; only the surrounding context grows. This lets us test whether retrieval remains sensitive to the same compositional edit as it occupies a smaller fraction of the caption. Long captions are constrained to remain within the 77-token context window of CLIP-style text encoders, so long-caption failures are not driven by truncation. Figure~\ref{fig:teaser} shows an example base/foil pair with captions at all three verbosity levels.

\paragraph{Scoring protocol.}\label{sec:scoring} CompoSET is scored as a $2\times2$ retrieval task. For each instance, the model scores four image-caption similarities: base image with base caption, base image with foil caption, foil image with base caption, and foil image with foil caption. Image-to-text (I2T) passes when each image assigns higher similarity to its matching caption than to the other caption. Text-to-image (T2I) passes when each caption assigns higher similarity to its matching image than to the other image. Group passes when both directions pass on the same instance. These metrics follow Winoground~\cite{thrush2022winoground}, with I2T/T2I naming from BiVLC~\cite{miranda2024bivlc}.

Formally, for an instance with base image $i_b$, foil image $i_f$, base caption $c_b$, and foil caption $c_f$, and writing $s(c, i)$ for the cosine similarity between caption $c$ and image $i$:
\begin{align}
\text{I2T}   &= \mathds{1}\!\left[\, s(c_b,i_b) > s(c_f,i_b) \;\land\; s(c_f,i_f) > s(c_b,i_f) \,\right] \\
\text{T2I}   &= \mathds{1}\!\left[\, s(c_b,i_b) > s(c_b,i_f) \;\land\; s(c_f,i_f) > s(c_f,i_b) \,\right] \\
\text{Group} &= \text{I2T} \cdot \text{T2I}
\end{align}
The benchmark score is the average pass rate over instances. There is no partial credit. Chance is 25\% for I2T and T2I, and 16.67\% for Group. Each instance has four similarity scores: two for the correct image-caption pairings and two for the incorrect ones. Group passes exactly when both correct pairings outscore both incorrect ones; for four random scores, the chance that the top two land on the two correct pairings is 1 in 6. Formally, let $a = s(c_b, i_b)$ and $d = s(c_f, i_f)$ be the two matching similarities and $b = s(c_b, i_f)$, $c = s(c_f, i_b)$ the two mismatched ones. By definition $\text{I2T} = \mathds{1}[a > c \wedge d > b]$ and $\text{T2I} = \mathds{1}[a > b \wedge d > c]$, so $\text{Group} = \mathds{1}[\min(a,d) > \max(b,c)]$: Group passes iff the two matching scores are the top two of the four. Under the null the four scores are exchangeable, so each of the $4! = 24$ rankings is equally likely, and exactly $2! \times 2! = 4$ place $\{a,d\}$ above $\{b,c\}$. Hence $P(\mathrm{Group}) = 4/24 = 1/6$. Each direction alone passes with probability $1/4$ (two comparisons of disjoint score pairs, each won with probability $1/2$); the product $1/16$ would require independence, but the directions share the same four scores and are positively dependent.

Generative VLMs require different scoring primitives, such as caption log-likelihood or prompted pairwise selection. We report them separately in App.~\ref{app:gen-vlm} and do not compare them head-to-head with dual encoders.

\paragraph{Construction and quality control.}\label{sec:pipeline}\label{sec:qc} CompoSET is built with a four-stage data engine using a frontier captioning model and a frontier image-generation model. First, the captioning model writes a structured specification for a naturalistic base scene. Second, it proposes taxonomy-grounded edits to objects present in that scene. Third, the image model renders each foil from the base scene using a custom single-edit prompt designed to apply only the named edit while preserving the rest of the scene. Fourth, the captioning model writes paired short, medium, and long captions for each accepted base/foil instance. Prompt templates and full construction details are in App.~\ref{app:construction}.

Every candidate instance is manually reviewed by an author of this paper. The review checks four criteria: the named object count is preserved; no unrelated object is added, removed, moved, or fully occluded; the visual edit matches the caption edit; and ambiguous captions disambiguate the target by location or appearance. Medium and long captions are then reviewed for fidelity to the short edit phrase. Candidates failing these criteria are discarded. Overall, 22.4\% of attempted foils are removed (511 of 2{,}287 attempts), with the highest discard rates for \texttt{modify\_presence} (41\%), \texttt{swap\_material} (35\%), and \texttt{modify\_transparency} (31\%). Full QC details are in App.~\ref{app:construction}.

\section{Dataset Analysis}
\label{sec:analysis}

\begin{figure}[t]
\centering
\includegraphics[width=\textwidth]{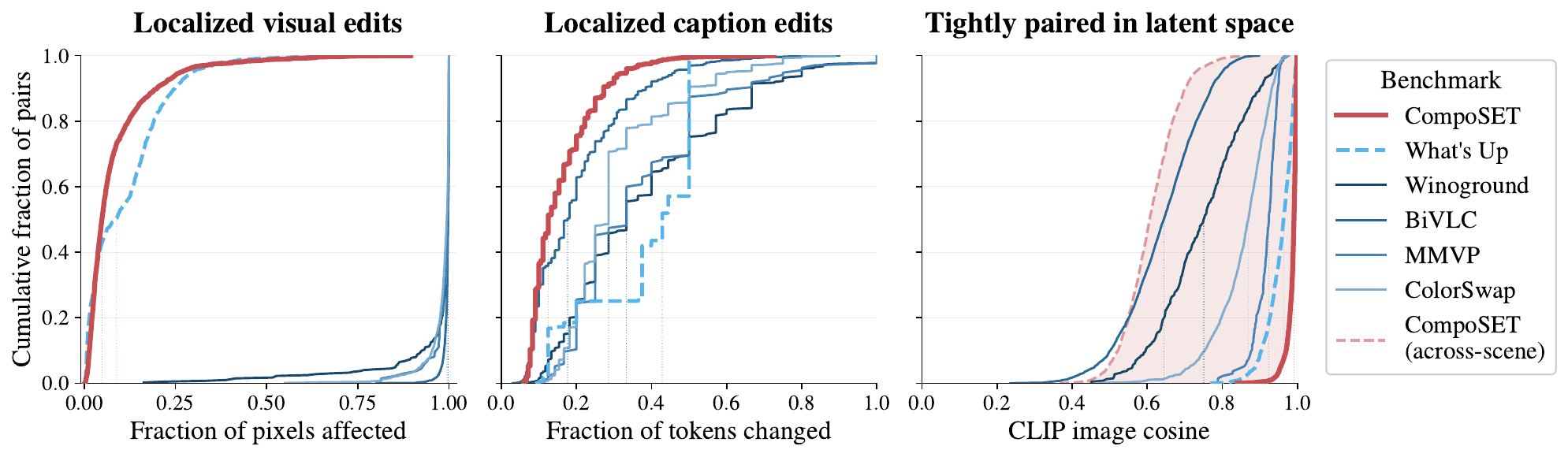}
\caption{\textbf{CompoSET base/foil pairs are more controlled than every prior benchmark on three independent axes.} Each panel plots the empirical cumulative distribution function (ECDF) of one per-pair measurement across compositionality benchmarks. {\bf (Left)} fraction of pixels affected per (base, foil) pair (Sec.\ref{sec:stats-image-dev}). {\bf (Middle)} fraction of caption tokens changed at the medium verbosity tier (Sec.\ref{sec:stats-text-dev}). {\bf (Right)} within-pair CLIP image cosine (Sec.\ref{sec:stats-cosine}). The controlled-edit side is left for the first two panels (smaller edits) and right for the third (tighter pairs); on all three axes CompoSET sits well to that side of every prior benchmark.}
\label{fig:analysis-trio}
\end{figure}

Before evaluating models, we verify that CompoSET has the properties required for controlled compositional evaluation. The analyses below test four claims: base and foil images differ by localized edits rather than whole-scene variation; scenes remain naturalistic and multi-object; caption verbosity varies context while preserving the edit phrase; and base/foil pairs are much tighter than random scene pairs in embedding space. We compare CompoSET against different-scene benchmarks, where base and foil are unrelated images that share a topic but not a setting (Winoground~\cite{thrush2022winoground}, BiVLC~\cite{miranda2024bivlc}, ColorSwap~\cite{burapacheep2024colorswap}, MMVP~\cite{tong2024mmvp}, NaturalBench~\cite{li2024naturalbench}), and same-scene benchmarks, where base and foil share a visual setup with one element edited (What's Up~\cite{kamath2023whatsup}, ColorBench-robust~\cite{liang2025colorbench}; SPEC~\cite{peng2024spec} and EqBen~\cite{wang2023eqben} are included where image-deviation statistics are available). Figure~\ref{fig:analysis-trio} summarizes the three controlled-axis claims at a glance: CompoSET pairs differ in fewer pixels, share more caption tokens, and are tighter in CLIP image space than every prior benchmark.

\paragraph{Controlled visual edits.}\label{sec:stats-image-dev} CompoSET base/foil images differ by a localized edit, not whole-scene variation (Figure~\ref{fig:analysis-trio}, left). The mean fraction of pixels changed between base and foil is 0.083 on CompoSET (median 0.048), compared with $\sim$0.97 on different-scene benchmarks (Winoground, BiVLC, ColorSwap, MMVP, NaturalBench). Same-scene peers are closer in pixel space: What's Up changes 0.111 of pixels on average, and ColorBench-robust changes 0.664. This places CompoSET in the same-scene regime, while extending beyond existing same-scene benchmarks in scene complexity, edit coverage, and foils per base. What's Up is similarly localized but limited to two-object scenes and four spatial prepositions; CompoSET spans 16 edit types across naturalistic multi-object scenes. Full image-deviation metrics and per-edit-type breakdowns are in App.~\ref{app:image-deviation}.

\paragraph{Naturalistic multi-object scenes.}\label{sec:stats-objcount} CompoSET is controlled without being object-centric. OWL-ViT v2~\cite{minderer2023owlv2} detections at threshold 0.3 over 116 generic categories find a mean of 20.7 objects per CompoSET scene (median 16, IQR 11.2--26.8, range 3--66, measured on a 50-scene subset), compared with 5.1 on Winoground, 6.1 on BiVLC, and 6.3 on SugarCrepe. The contrast holds across the distribution: CompoSET's 25th percentile (11.2 objects per scene) exceeds the 75th percentile of every prior benchmark we measured. People are not the cause; removing people shifts all benchmarks by 0.9--2.6 objects without changing the rank order. Multi-object scenes make the task genuinely compositional: the question is not only whether an attribute appears, but whether it is assigned to the correct entity. Full distributions and detector ablations are in App.~\ref{app:objcount}.

\paragraph{Controlled caption verbosity.}\label{sec:stats-text-dev}\label{sec:stats-descriptive} Caption verbosity is a controlled axis in CompoSET: the edit phrase stays fixed while surrounding scene context grows (Figure~\ref{fig:analysis-trio}, middle). Caption lengths increase from a median of 4 tokens at short verbosity to 7 at medium and 25 at long. As captions move from short to medium to long, the fraction of changed tokens drops from 0.38 to 0.16 to 0.09. Sentence-embedding similarity rises in parallel: MPNet cosine increases from 0.85 to 0.92 to 0.97. Thus, longer captions are dominated by shared context, and the base/foil difference concentrates in the same edit phrase. Prior benchmarks do not provide this axis. What's Up, the closest benchmark in image edit fraction, changes 0.38 of caption tokens at its only caption level, matching CompoSET-short but remaining well above CompoSET-medium and CompoSET-long. Full per-verbosity and per-edit-type analyses are in App.~\ref{app:text-deviation}.

\paragraph{Same-scene tightness in embedding space.}\label{sec:stats-cosine} CompoSET base/foil pairs are tight in CLIP image space; prior benchmark pairs are often closer to random scene pairs (Figure~\ref{fig:analysis-trio}, right). The CLIP image cosine between a CompoSET base and its foil is 0.986 (median 0.991, IQR 0.983--0.995), while the cosine between two unrelated CompoSET base scenes is 0.609. The within-vs-across gap is 0.38, the widest among the benchmarks we measured. Across prior benchmarks, the gap ranges from 0.15 to 0.33 (Winoground 0.17, BiVLC 0.15, ColorSwap 0.26, MMVP 0.31, NaturalBench 0.24, What's Up 0.17, ColorBench-robust 0.33). On Winoground and BiVLC, the intended compositional pair is only slightly tighter than two random images from the benchmark, making compositional reasoning difficult to disentangle from global scene differences. The full per-benchmark within-vs-across distributions are summarized in Figure~\ref{fig:analysis-trio}, right.

\section{Experiments}
\label{sec:experiments}
\label{sec:experiments-models}

We evaluate CompoSET on 22 dual-encoder VLMs and compare against nine prior compositionality benchmarks. Dual encoders are the primary target: they score image-caption matches by cosine similarity in a shared embedding space, the same scoring primitive used by CompoSET. We also report open-weight generative VLMs and frontier API models separately, because they use different scoring primitives and are not directly comparable to dual encoders.

\paragraph{Models and scoring.} The dual-encoder pool spans four research directions: contrastive baselines (CLIP~\cite{radford2021clip}; the open LAION re-implementation~\cite{schuhmann2022laion}; SigLIP~\cite{zhai2023siglip}; SigLIP-2~\cite{tschannen2025siglip2}), text-side hard-negative training (NegCLIP~\cite{yuksekgonul2023bow}; CLoVe~\cite{castro2024clove}), caption-enrichment pretraining (DreamLIP~\cite{zheng2024dreamlip}), and data-curation recipes (DFN~\cite{fang2024dfn}; MetaCLIP-2~\cite{chuang2025metaclip2}; EVA-02-CLIP~\cite{sun2023evaclip,fang2024eva02}). The pool also includes Jina-embeddings-v4~\cite{jina2025embeddingsv4}, a single-vector embedding model built on a generative VLM backbone. For dual encoders, we use the scoring protocol from \S\ref{sec:scoring}. For open-weight generative VLMs, candidate captions are scored by length-normalized caption log-likelihood under the image. For frontier API models, we use prompted pairwise selection with option-order counterbalancing to control position bias. These generative and frontier results are reported as caveated slices rather than head-to-head comparisons.

\label{sec:results}

\begin{table}[t]
\centering
\caption{\textbf{Every dual encoder is at or below chance on CompoSET, while most are above chance on prior benchmarks.} The exceptions are Winoground and What's Up, where every model is also at or below chance. Compositionality results across 10 benchmarks for 14 representative dual encoders, grouped by contribution axis (\S\ref{sec:experiments-models}). Best-in-pool per column in \textbf{bold}.}
\label{tab:main-results}
\small
\setlength{\tabcolsep}{3pt}
\resizebox{\textwidth}{!}{%
\begin{tabular}{l>{\columncolor{black!7}\centering\arraybackslash}p{30pt}>{\centering\arraybackslash}p{30pt}>{\centering\arraybackslash}p{30pt}>{\centering\arraybackslash}p{30pt}>{\centering\arraybackslash}p{30pt}>{\centering\arraybackslash}p{30pt}>{\centering\arraybackslash}p{30pt}>{\centering\arraybackslash}p{30pt}>{\centering\arraybackslash}p{30pt}>{\centering\arraybackslash}p{30pt}}
\toprule
Model & \makecell{\textbf{Compo}\\\textbf{SET}} & \makecell{Wino\\ground} & BiVLC & \makecell{What's\footnotemark\\Up} & \makecell{Color\\Swap} & MMVP & \makecell{Natural\\Bench} & \makecell{Color\\Bench} & \makecell{Sugar\\Crepe} & \makecell{Sugar\\Crepe\textsuperscript{++}} \\
\midrule
\emph{Chance} & \emph{16.7} & \emph{16.7} & \emph{16.7} & \emph{16.7} & \emph{16.7} & \emph{16.7} & \emph{50.0} & \emph{50.0} & \emph{50.0} & \emph{50.0} \\
\midrule
\multicolumn{11}{l}{\emph{Contrastive baselines}} \\
CLIP B/32 & 4.4 & 7.5 & 47.2 & 2.2 & 8.7 & 8.1 & 68.1 & 46.7 & 78.6 & 65.5 \\
CLIP B/16 LAION-2B & 7.8 & 8.2 & 53.3 & 1.1 & 19.3 & 12.6 & 70.7 & 50.6 & 81.3 & 70.1 \\
CLIP H/14 LAION-2B & 8.4 & 10.5 & 56.9 & 0.5 & 32.0 & 18.5 & 73.5 & 55.4 & 82.5 & 70.9 \\
SigLIP B/16 & 10.2 & 10.2 & 59.6 & 1.6 & 30.7 & 23.0 & 74.7 & 55.4 & 81.3 & 70.3 \\
SigLIP SO400M/14 & 13.5 & 12.2 & \textbf{60.0} & 2.8 & 37.0 & 28.9 & 77.2 & 56.9 & 83.6 & 73.7 \\
SigLIP-2 B/16 & 11.3 & 10.8 & 57.6 & 2.1 & 34.7 & 23.7 & 75.9 & 56.0 & 84.1 & 75.4 \\
SigLIP-2 L/16-384 & 13.6 & 14.0 & 52.8 & 2.3 & 44.3 & 23.0 & 78.8 & \textbf{61.4} & 83.3 & 74.6 \\
SigLIP-2 SO400M/14 & \textbf{13.9} & \textbf{15.2} & 49.5 & 4.2 & 39.7 & \textbf{31.1} & \textbf{79.4} & 54.3 & 84.3 & \textbf{75.8} \\
\midrule
\multicolumn{11}{l}{\emph{Hard-negative compositional training}} \\
NegCLIP B/32 & 7.0 & 8.0 & 55.3 & 4.1 & 14.0 & 12.6 & 72.1 & 43.0 & 83.4 & 67.4 \\
CLoVe B/32 & 5.5 & 4.8 & 48.3 & 0.8 & 15.0 & 11.1 & 67.8 & 45.8 & \textbf{84.9} & 56.1 \\
\midrule
\multicolumn{11}{l}{\emph{Caption-enrichment pretraining}} \\
DreamLIP B/16 (30M) & 13.7 & 11.5 & 53.9 & \textbf{5.4} & \textbf{61.0} & 16.3 & 74.4 & 48.6 & 81.5 & 66.2 \\
\midrule
\multicolumn{11}{l}{\emph{Data-curation recipes}} \\
DFN H/14 & 10.8 & 11.0 & 59.4 & 3.5 & 45.0 & 30.4 & 75.8 & 57.5 & 84.0 & 74.6 \\
EVA-02 L/14 & 8.6 & 10.2 & 55.2 & 2.0 & 12.7 & 23.0 & 73.3 & 53.4 & 80.9 & 70.8 \\
MetaCLIP-2 H/14 & 8.9 & 11.0 & 48.3 & 1.6 & 20.7 & 17.0 & 72.9 & 53.2 & 80.2 & 69.3 \\
\bottomrule
\end{tabular}%
}
\end{table}
\footnotetext{What's Up scores reflect a fixed preposition bias in dual-encoder models. For example, CLIP-B/32 correctly identifies `on' as the correct caption 92.2\% of the time, but only picks `under' correctly 22.3\% of the time. Strict bidirectional Group exposes this by requiring both image-to-text and text-to-image directions to be correct on the same pair; see App.~\ref{tab:whatsup-per-prep}.}

\paragraph{Results.} Table~\ref{tab:main-results} reports 14 representative dual encoders; the full 22-model pool and confidence intervals are in App.~\ref{app:full-results}. The results support seven findings: controlled single-edit evaluation places every dual encoder at or below chance; scale improves performance only modestly; text-side hard-negative training does not transfer to controlled image-side foils; caption verbosity matters, with DreamLIP's dense-recaption pretraining providing the clearest robustness; matched modify/swap contrasts show attribute binding collapsing even where detection survives; many foils per scene separate scene-hard from edit-hard failures; and frontier VLMs recover more of the signal, but only with per-query scoring.

\textbf{Dual encoders perform at or below chance on CompoSET despite strong scores elsewhere.} The best CompoSET Group score is 13.9\% (SigLIP-2 SO400M/14, 95\% CI $[12.3, 15.5]$; scene-clustered bootstrap CI $[12.4, 15.5]$; $n{=}1{,}776$), below the 16.67\% chance level. DreamLIP B/16 is close behind at 13.7\%, despite training on substantially less data. The same SigLIP-2 SO400M scores 49.5\% on BiVLC, 39.7\% on ColorSwap, and 31.1\% on MMVP. The gap shows that CompoSET is not merely re-ranking existing benchmark leaders: models that appear compositionally competent under prior protocols fail when the image contrast is controlled. The failure is concentrated on Spatial, Quantity, and most Activity \& State edits, with non-trivial performance only on a handful of Identity \& Appearance edits (App.~\ref{app:per-edittype-per-verbosity}, Figure~\ref{fig:per-edittype-range}). A clustered bootstrap over the 80 base scenes ($B{=}10{,}000$; App.~\ref{app:repro}) widens the Wald intervals of Table~\ref{tab:app-full-results} by 27--71\%, and every one of the 22 upper bounds remains below the 16.67\% chance level (maximum 15.55\%).

\textbf{Scale and data improve scores, but do not solve the controlled task.} CompoSET Group rises from CLIP B/32 (4.4\%) to CLIP B/16 LAION-2B (7.8\%), SigLIP B/16 (10.2\%), and SigLIP-2 SO400M/14 (13.9\%). This is a meaningful gain, but the absolute level remains at or below chance. Scaling the dual-encoder recipe improves controlled compositional discrimination only modestly.

\textbf{Text-side hard negatives do not transfer to image-side foils.} NegCLIP B/32 scores 7.0\% on CompoSET, below CLIP B/16 LAION-2B at 7.8\%. NegCLIP fine-tunes CLIP on captions whose noun phrases are syntactically swapped, perturbing only the text side. CompoSET changes the image as well as the caption, so the failure mode is different: the model must preserve the corresponding visual alternative, not just prefer a better caption for a fixed image.

\textbf{Caption verbosity matters, and dense-recaption pretraining is the most robust dual-encoder axis.} Most cosine dual encoders lose 2--3 points of Group score from short to long captions on CompoSET: SigLIP-2 SO400M drops from 14.6\% to 12.4\%, SigLIP-2 L/16-384 from 14.7\% to 12.0\%, and CLIP B/16 LAION-2B from 9.1\% to 6.5\%. Since long captions stay within the 77-token cap of CLIP-style encoders by construction, the degradation reflects dilution of the fixed edit phrase rather than truncation. DreamLIP B/16, which recaptions each training image with a detailed MLLM caption and samples multiple short sub-captions per image as positives with subcaption-level alignment, is flat across verbosity levels (13.1\%, 14.2\%, 13.7\%) and tops the long-caption ranking. This is the main empirical payoff of the verbosity axis: most dual encoders lose the edit as context grows, while dense multi-positive caption supervision partially protects it.

\textbf{Swap edits isolate binding, and binding collapses even where detection survives.} A swap edit exchanges attribute values between two objects, so both values appear in both captions and attribute detection is uninformative: a bag-of-words matcher scores at chance on swaps. The matched modify edit probes the same attribute in the same scenes with detection left intact. The contrast is stark: \texttt{modify\_color} spans 17.2--52.4\% Group across the 22 encoders, while \texttt{swap\_color} sits at or below 5.8\% for 21 of them --- far below the 16.67\% chance level, meaning encoders systematically prefer the wrong assignment rather than guess. The modify-over-swap gap is positive for every encoder on both matched attributes (color $+15.9$ to $+36.3$; material $+1.7$ to $+14.4$, with \texttt{modify\_material} at 1.7--16.4\% against \texttt{swap\_material} at 0.0--4.3\%). Attribute detection therefore partially survives while attribute binding collapses. Winoground's swap-only construction already suggested binding failure~\cite{thrush2022winoground}; the controlled matched-attribute contrast within fixed scenes attributes it. The one exception is consistent with the caption-enrichment finding above: DreamLIP B/16 reaches 18.6\% on \texttt{swap\_color} (its own \texttt{modify\_color}: 52.4\%), suggesting dense-recaption, multi-positive pretraining confers some binding signal as well. Per-edit-type results for all models are in Table~\ref{tab:app-per-edittype}.

\textbf{Many foils per scene separate scene-hard from edit-hard failures.} With roughly 22 foils per base scene, CompoSET can ask a question one-foil-per-base benchmarks cannot: when a model fails, is the scene the obstacle, or the edit? For each model-scene pair, we label the bottom quartile of mean I2T margin \emph{scene-hard} (every edit in the scene is hard) and the top quartile of foil-to-foil margin variance \emph{edit-hard} (specific edits fail in an otherwise tractable scene). The two failure modes are nearly independent: 4.7\% of model-scene pairs carry both labels, against 6.25\% expected under independence. The decomposition is stable across models. Per-scene margin rankings correlate at mean Spearman $\rho=0.56$ across the 22 encoders, scene identity explains 53\% of per-scene margin variance (model identity: 10\%), and 17 of 80 scenes are scene-hard for a majority of models. Edit-hardness is not a relabeling of edit-type difficulty: removing per-edit-type means leaves the edit-hard ranking essentially unchanged ($\rho=0.95$). Most of the signal lives at the foil level --- 94.7\% of per-foil success-rate variance lies within scenes --- so a one-foil-per-base benchmark samples only the remaining 5.3\%. Full protocol and robustness checks are in App.~\ref{app:cosine-margin}.

\textbf{Frontier API models partially close the gap, but only at much higher per-pair inference cost.} We run a small-scale experiment on a stratified 200-instance subset, sized to bound API cost. Frontier models are scored by prompted pairwise selection: each $2\times2$ instance issues four calls (two I2T, two T2I) asking the model to choose between two candidate options. Frontier models systematically prefer a fixed option position regardless of which option is correct: Gemini Flash favors the first option (76.5\% T2I when the correct image is first vs.\ 30.0\% when second), Haiku 4.5 favors the second. We counterbalance every call by issuing it twice with the option order swapped and averaging the binary correctness. Under this protocol, Claude Sonnet 4.6 reaches 42.5\% Group on CompoSET, about $3\times$ the best dual-encoder ceiling and well above chance; Gemini 2.5 Flash and Claude Haiku 4.5 reach 12.0\% and 13.0\%. Pairwise selection costs four API calls per instance, so frontier compositional capability is not pre-indexable for scalable retrieval. Open-weight generative VLMs scored by caption log-likelihood are reported as a separately caveated slice in App.~\ref{app:gen-vlm}; their scoring primitive is not directly comparable to dual-encoder cosine.

\section{Limitations and Discussion}
\label{sec:discussion}

\textbf{Quality over scale.} CompoSET contains 80 scenes and 1{,}776 instances: smaller than SugarCrepe, larger than Winoground or What's Up. This scale reflects the cost of preserving the benchmark's central property. Same-scene single-edit foils require human review, and the image-deviation analysis (§\ref{sec:stats-image-dev}) shows the result: base/foil images differ by 0.083 of pixels on average, compared with $\sim$0.97 on different-scene benchmarks. Scaling to tens of thousands of instances without comparable review would risk reintroducing the scene-level drift CompoSET is designed to control.

\textbf{One edit type is structurally rare.} \texttt{swap\_role} requires at least two distinct agents in a scene, and many CompoSET scenes have one subject or none. After filtering, \texttt{swap\_role} has $n=45$ instances, below our preferred $n=50$ per-edit reporting floor. We keep it in the overall benchmark totals, but report per-edit results with a confidence-interval caveat in App.~\ref{app:per-edittype-per-verbosity}. Forcing more \texttt{swap\_role} cases would require biasing scene generation toward multi-agent interactions, reducing scene diversity.

\textbf{Single-rater review.} One author reviewed every candidate instance against the explicit four-criterion rubric in §\ref{sec:qc}; 22.4\% of candidates were discarded. A separate drift audit, in which Gemini 2.5 Flash flags candidate changes and each flag is verified manually, checks the no-unintended-change criterion and finds 0.67\% residual drift on $N{=}300$ pairs (App.~\ref{app:repro}). No inter-rater reliability statistic is available.

\textbf{Some edits are less pixel-local than others.} The mean pixel-change fraction of 0.083 hides variation across edit types. \texttt{swap\_role}, \texttt{modify\_pose}, and \texttt{modify\_action} each fall between 0.21 and 0.31, three to four times the dataset mean (App.~\ref{app:image-deviation}). These edits perturb posture, contact geometry, or agent roles, so the image generator often regenerates a larger region than it does for color or pattern changes. They remain far below the typical Winoground or BiVLC pair, but conclusions about these edit types should be read with this higher-variance caveat.

\textbf{What CompoSET does not test.} CompoSET uses model-synthesized images, English captions, and a single frontier image generator. It tests one edit at a time in static images; multi-edit composition and video compositionality remain open. The taxonomy deliberately excludes categories that are hard to control cleanly or ethically in synthetic imagery: age, gender, emotion, ethnicity, scene-level changes such as lighting or weather, absolute size, aesthetics, and ordinal position. Finally, CompoSET is a retrieval benchmark: it measures image-caption matching behavior, not internal mechanisms such as attention, probing directions, or causal circuits.

\paragraph{Ethics and responsible use.}
CompoSET is released under CC-BY-4.0. All images are model-synthesized; no real-person data, web-scraped content, or sensitive material is involved. Intended use is evaluation only; we do not provide a training split. Full license, consent, sensitive-content handling, and intended-use disclosure appear in Appendix~\ref{app:datasheet}, alongside the Datasheet.

\section{Conclusion}
\label{sec:conclusion}

We introduced CompoSET, a controlled single-edit benchmark for vision-language compositionality that is designed to test experimentally the properties of VLMs while avoiding the confounds that are inherent in current benchmarks. By pairing each base image-caption with a foil that differs by exactly one semantic edit while keeping the rest of the scene fixed, CompoSET isolates compositional discrimination from scene-level confounds. The construction holds 92\% of pixels fixed, varies one of 16 edit types, and provides $\sim$22 foils per base scene across three caption densities.

Across 22 dual-encoder VLMs, every model performs at or below chance on CompoSET, including SigLIP-2 SO400M, which scores 49.5\% on BiVLC and 75.8\% on SugarCrepe++. Thus, recently reported progress in dual-encoder compositionality is likely an artifact of benchmark design. With confounds removed, the field has a clear baseline to improve against.

Frontier VLMs scored via prompted pairwise selection reach 42.5\% Group on a 200-instance subset, three times the dual-encoder ceiling. Compositional capability thus exists in current systems but only at per-query inference cost — incompatible with the indexed-retrieval workloads dual-encoders serve.

CompoSET tests single edits in static images with English captions. Multi-edit composition, video compositionality, non-English text, and architectural changes that move beyond single-vector matching are natural next steps. CompoSET is evaluation-only by design; whether the controlled single-edit recipe can also inform training data — and whether models trained on such data improve under the same controlled protocol — is the natural follow-up question to be investigated.

\bibliographystyle{plain}
\bibliography{references}

\appendix

\section{Datasheet \& Responsible AI}
\label{app:datasheet}

This appendix follows the structure of \emph{Datasheets for Datasets}~\cite{gebru2021datasheets}, with a Responsible AI disclosure (license, consent, sensitive content).

\subsection{Datasheet for Datasets}

\paragraph{Motivation.} CompoSET was created to measure compositional understanding in vision-language models under controlled single-edit conditions. Existing compositionality benchmarks pair structurally dissimilar images, so model performance can reflect global image differences rather than the targeted compositional contrast (§\ref{sec:intro}). CompoSET addresses this gap: every base/foil image pair shares a scene and differs by exactly one labeled edit, so performance is interpretable as compositional discrimination on a single named factor. The dataset was created by the authors of this paper.

\paragraph{Composition.} An \emph{instance} is a tuple of two images (base, foil) and three matched caption pairs at three verbosity tiers (short, medium, long), plus an edit-type label drawn from a 16-category taxonomy across four semantic groups (Identity \& Appearance, Quantity, Spatial Relations, Activity \& State). The release contains 1{,}776 instances built from 80 base scenes, with at least 15 and on average $\sim$22 foils per base. Per-edit-type counts are listed in Table~\ref{tab:taxonomy}. Images are PNG, 1024\ding{53}1024. Captions are ASCII English. Each instance carries an edit-type label and a per-scene scene-id but no other annotations.

\paragraph{Collection process.} All images are model-synthesized; no real-world images were collected. The construction pipeline (§\ref{sec:pipeline}, App.~\ref{app:construction}) uses a frontier captioning model (Anthropic Claude Sonnet 4.6) for scene specification, edit enumeration, and caption writing, and a frontier image-generation model (Google Gemini 2.5 Flash Image) for base and foil rendering. A custom single-edit prompt steers the image model to apply only the named edit while preserving the rest of the scene; details in App.~\ref{app:construction}. No human subjects were recruited and no PII was collected.

\paragraph{Preprocessing / labeling.} Every instance underwent manual quality-control review by an author of this paper, applied across three phases: (i) base-caption review before any foil is generated; (ii) per-instance review of base/foil images and short captions against four criteria (count preserved, no other object changed, visual edit matches textual edit, disambiguating caption when ambiguous); (iii) review of medium and long captions for fidelity to the short edit phrase. Phase-two failure rate is 22.4\% (511 of 2{,}287 attempts discarded). Discard rates by edit type are highest for \emph{modify\_presence} (41\%), \emph{swap\_material} (35\%), and \emph{modify\_transparency} (31\%) (see App.~\ref{app:construction}).

\paragraph{Uses.} CompoSET is intended for evaluation of vision-language model compositional understanding under controlled single-edit conditions. The primary intended use is benchmarking dual-encoder VLMs via bidirectional image-caption matching (Winoground-style $2{\times}2$ scoring); a generative-VLM slice using caption log-likelihood or prompted pairwise selection is also supported (App.~\ref{app:gen-vlm}). The dataset is \emph{not} a training resource: we provide no train split, and the small scale relative to web-scraped pretraining datasets makes it unsuitable for that purpose. The dataset should not be used to make claims about open-ended visual reasoning beyond the 16 covered edit types.

\paragraph{Distribution.} CompoSET is released on Hugging Face at \texttt{huggingface.co/datasets/CompoSET/CompoSET} under CC-BY-4.0 (with attribution to the original authors). The release includes (a) the 1{,}776 base/foil image pairs as PNG files; (b) a \texttt{variations.parquet} table mapping each instance to its edit type, scene id, and three caption pairs; (c) a README with citation instructions and a quickstart pointer to the evaluation code.

\paragraph{Maintenance.} The dataset is versioned; the released v1.0 freeze hash is \texttt{662230d}. We do not plan to extend the dataset beyond v1.0 under this paper, but errata (typo fixes, image regenerations of QC-flagged instances) may be released as v1.0.x with changelog entries on the HF dataset card. Issues can be filed on the HF dataset repo.

\subsection{Responsible AI Disclosure}

\paragraph{License.} The released dataset is licensed CC-BY-4.0. The companion evaluation code is licensed MIT and is available at \url{https://github.com/laure-delisle/CompoSET}.

\paragraph{Consent.} No human subjects participated in the creation or use of CompoSET. All images are model-synthesized from scene specifications written by the authors and a captioning model; no images were scraped from the web; no real persons or private spaces are depicted.

\paragraph{Sensitive content.} CompoSET contains naturalistic images of everyday objects and scenes. People appear in some scenes (32 of 80 contain at least one human figure); these are model-rendered humans, not real individuals, and they are referred to in captions only as ``a person'' (no gendered pronouns; see §\ref{sec:taxonomy}). The benchmark deliberately excludes age, gender, emotion, ethnicity, and other person-centric attributes. No medical, legal, or otherwise sensitive content is included. Manual QC (§\ref{sec:qc}) acted as a content filter: any image with unintentional artifacts, partial nudity, or otherwise objectionable content was discarded before release.

\paragraph{Broader impact.} The intended positive impact is more reliable measurement of compositional capability in vision-language models, surfacing benchmark-design confounds that have inflated reported progress. Negative impacts: the benchmark could be optimized against, and a model fine-tuned on CompoSET-style hard negatives might overfit to its specific edit taxonomy. We mitigate by releasing CompoSET as evaluation-only, providing no train split, and explicitly disclosing the taxonomy boundaries.

\section{Dataset Construction \& QC}
\label{app:construction}

CompoSET pairs are built by interleaving caption generation (Claude Sonnet 4.6) and image generation (Gemini 2.5 Flash Image, ``Nano Banana 2''). We organize the protocol by stage: \emph{base scene} (\S\ref{app:construction-base}) and \emph{foils} (\S\ref{app:construction-foils}), then a \emph{verbosity build} stage that expands kept short captions into medium and long tiers (\S\ref{app:construction-verbosity}). Each prompt template below is reproduced verbatim from \texttt{prompts/} in the released repository.

\subsection{Base scene}
\label{app:construction-base}

\textbf{Step 1: base scene caption.} Given scene parameters (setting, number of subjects, framing), the captioning model writes a structured base specification --- a one-sentence base caption plus per-region detail sentences --- rich in swappable attributes so that downstream edits can target real objects.

\promptfile{prompts/caption_1_generate_base.txt}

\textbf{Step 2: base image render.} The structured base caption drives a single image-generation call.

\promptfile{prompts/image_1_generate_base.txt}

\subsection{Foils}
\label{app:construction-foils}

\textbf{Step 3: per-instance edits to the base caption.} Given the rendered base image plus the original spec, the captioning model writes per-instance base captions: each one is anchored on the object whose attribute the foil edits, so two instances built from the same scene have different base captions, each phrased to make the relevant object salient.

\promptfile{prompts/caption_2_edit_from_image.txt}

\textbf{Step 4: foil-caption enumeration from the edit taxonomy.} Given the rendered base image and the 16-type edit taxonomy, the captioning model emits a set of $\sim$22 plausible single-edit foils per scene. Each foil is a \texttt{(edit\_type, base\_attribute, foil\_attribute, target\_object, base\_caption, foil\_caption)} tuple.

\promptfile{prompts/caption_3_generate_variations.txt}

\textbf{Step 5: self-check pass.} A second pass over the candidate foils drops cases where the edit cannot be applied to the scene, the target object is ambiguous, or the attribute change is implausible. Failed candidates are removed before any image is generated.

\promptfile{prompts/caption_4_verify_variations.txt}

\textbf{Step 6: foil image render.}
\label{app:imagegen}
The image model receives the base image plus base/foil captions and is instructed to change \emph{only} the targeted edit while preserving everything else. The aggressive object-preservation language is the primary lever that distinguishes CompoSET foils from off-the-shelf image-edit prompts and bounds drift to 0.67\% post-verification (App.~\ref{app:repro}).

\promptfile{prompts/image_2_apply_variation.txt}

\subsection{Verbosity build}
\label{app:construction-verbosity}

\textbf{Step 7: expand short to medium and long.} Given an accepted base/foil instance with a short caption (a noun phrase containing the edit attribute), the captioning model produces a \emph{medium} caption (one object-level anchor added) and a \emph{long} caption (full-scene description with the edit phrase fixed). The long tier is constrained to 77 BPE tokens (the CLIP text-encoder cap) so the long tier is not biased by truncation.

\promptfile{prompts/caption_6_build_levels.txt}

A bias-mitigation variant of each captioning prompt (\texttt{*.bias.txt}) is run in parallel as a sanity check; outputs from the two variants are reconciled by the verification pass.

\subsection{QC Protocol}

\begin{figure}[ht]
\centering
\includegraphics[width=\textwidth]{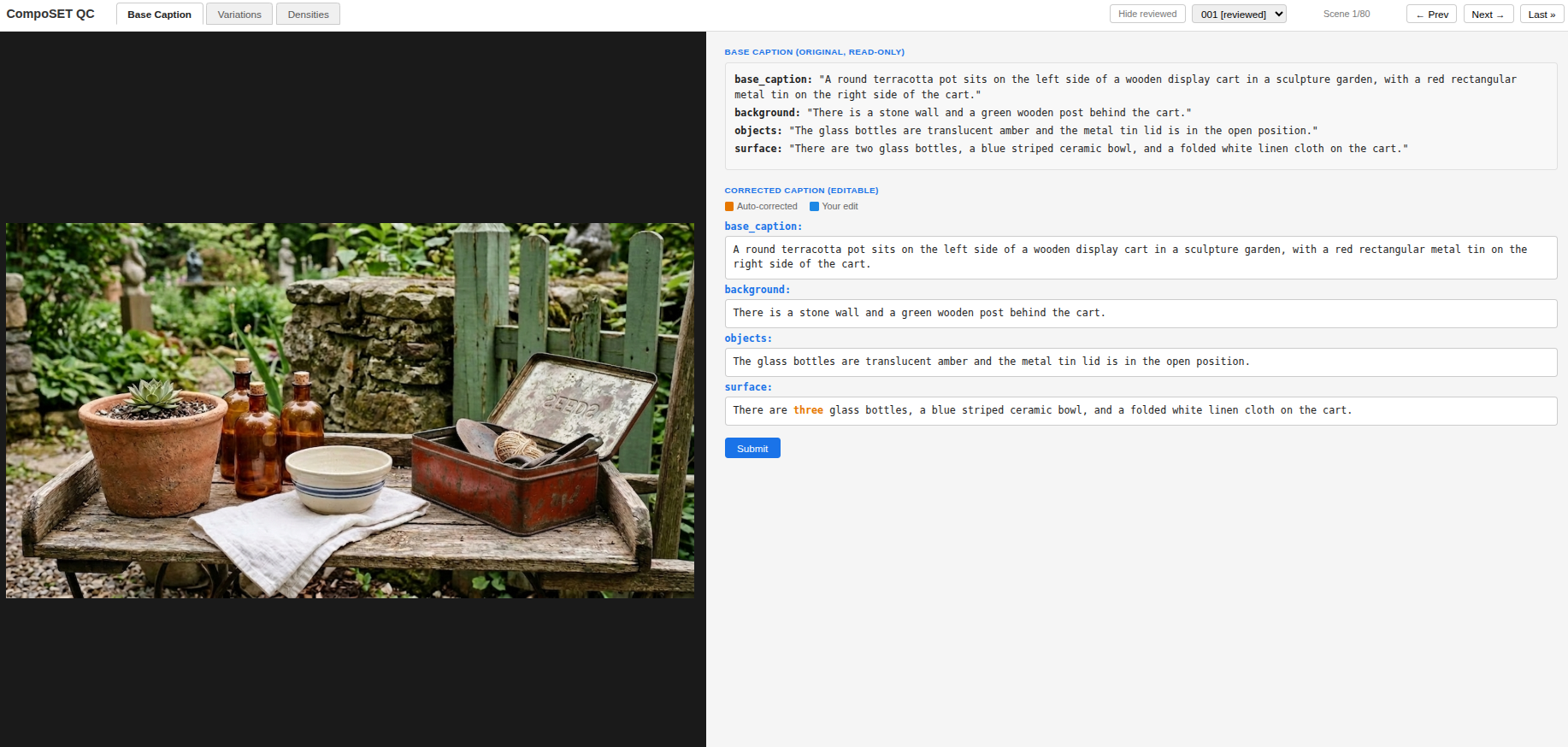}
\caption{Phase 1 (base review). Rater reviews the base caption against the rendered base image before any foils are generated; failures trigger scene re-generation.}
\label{fig:qc-base}
\end{figure}

\begin{figure}[H]
\centering
\includegraphics[width=\textwidth]{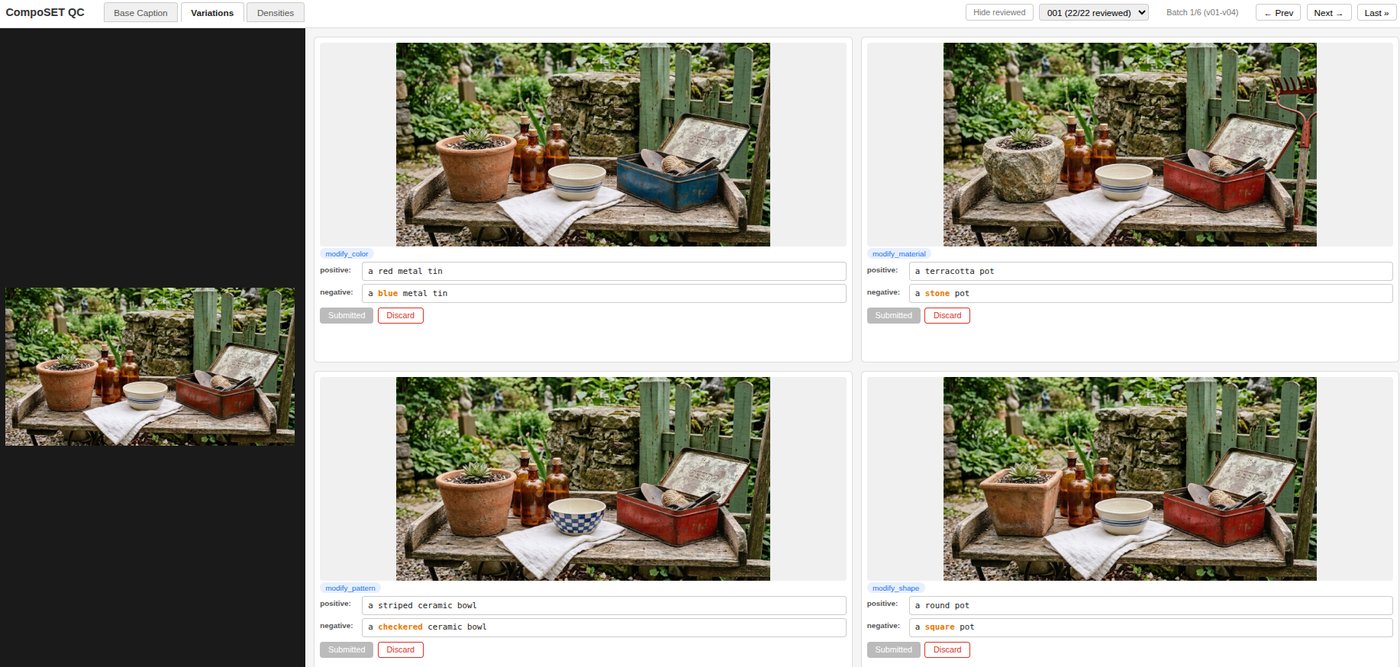}
\caption{Phase 2 (instance review). Rater compares each base/foil pair against the four-criterion rubric (count preserved, no other object changed, visual edit matches textual edit, caption disambiguates when ambiguous). Failures discard the instance.}
\label{fig:qc-variations}
\end{figure}

\begin{figure}[H]
\centering
\includegraphics[width=\textwidth]{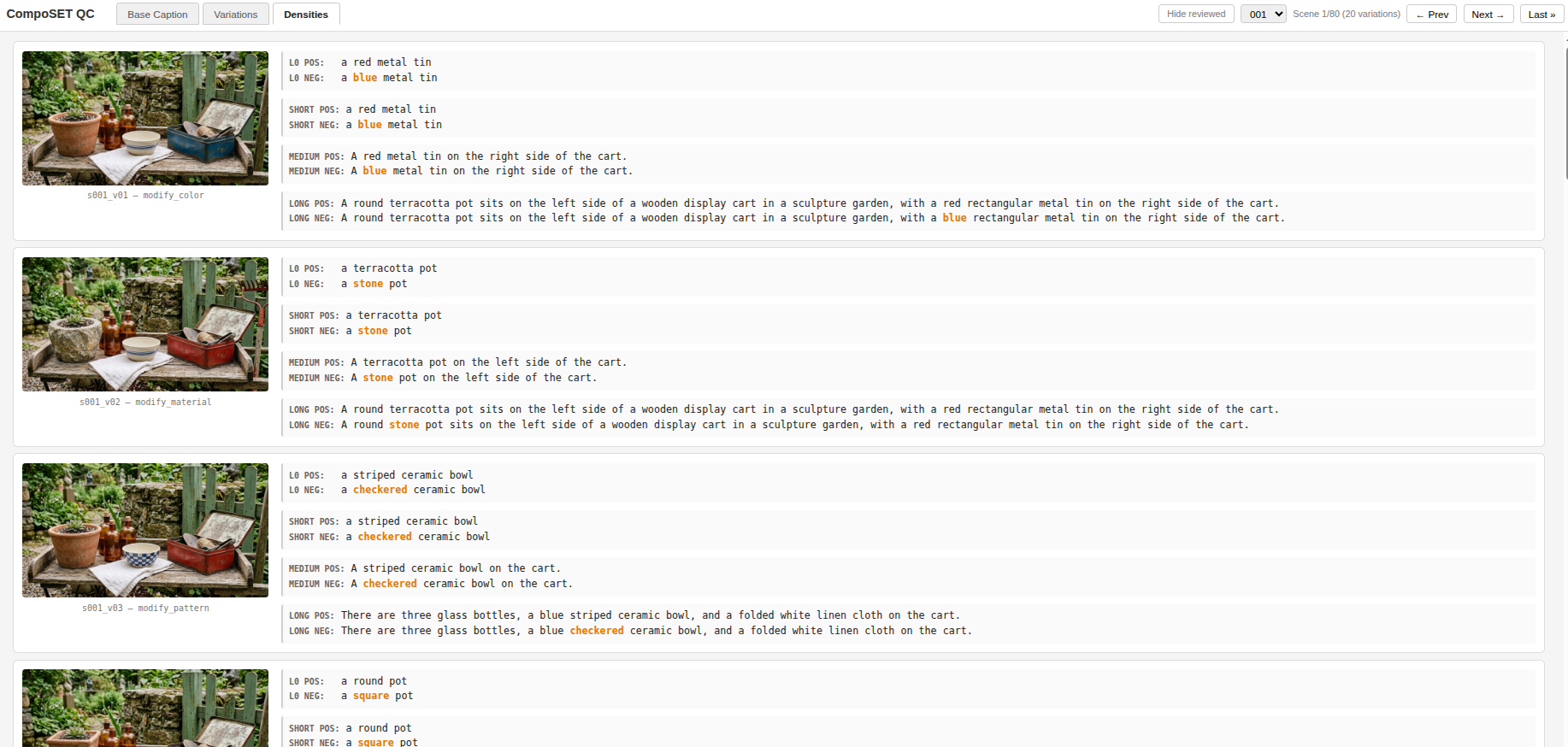}
\caption{Phase 3 (verbosity review). For each kept instance, rater verifies that the edit phrase is preserved across short, medium, and long captions while surrounding scene context grows without rewording.}
\label{fig:qc-densities}
\end{figure}

A single author of this paper reviewed every scene in three phases. Figures~\ref{fig:qc-base}--\ref{fig:qc-densities} show the QC interface used at each phase.

\textbf{Phase 1 (base review).} The base caption is reviewed before any foil is generated; ambiguous anchors and missing-object captions are corrected at this point. A failure here invalidates the entire scene and triggers re-generation.

\textbf{Phase 2 (instance review).} Each instance's images and short captions are reviewed against four criteria: (i) the count of the named object is preserved across base and foil; (ii) no other object is added, removed, moved, or fully occluded; (iii) the visual edit matches the textual edit; (iv) when two objects could plausibly match the edit description, the caption disambiguates by location or appearance. Failure on any criterion discards the instance. Phase-two discard rate is 22.4\% (511 of 2{,}287 attempts), highest on \texttt{modify\_presence} (41\%), \texttt{swap\_material} (35\%), and \texttt{modify\_transparency} (31\%) --- edits where the generator most often introduces secondary changes.

\textbf{Phase 3 (verbosity review).} Medium and long captions for each kept instance are reviewed for fidelity to the short edit phrase. The edit phrase must appear unchanged across the three tiers; surrounding context grows around it without rewording.

\textbf{Single rater.} One author reviewed every instance against the four-criterion rubric above; 22.4\% of candidates were discarded. The generator drift audit (App.~\ref{app:repro}) separately checks criterion (ii), no unintended change, and finds 0.67\% residual drift on $N{=}300$ pairs. No inter-rater reliability statistic is available.

\section{Reproducibility Artifacts}
\label{app:repro}

\paragraph{Confidence intervals.} All scores and rates reported in this paper are binomial proportions over a fixed test set: $\hat{p} = k/n$, where $k$ is the number of successes and $n$ is the number of pairs. We report 95\% confidence intervals; the choice between the two standard formulas depends on the regime.

For scores in the mid-range (roughly $0.1 \le \hat{p} \le 0.9$), we use the standard normal (Wald) approximation,
\[
\hat{p} \pm 1.96 \sqrt{\hat{p}(1-\hat{p})/n},
\]
clamped to $[0, 1]$ when the interval extends outside the unit range. Wald is symmetric and reads cleanly inline.

For rates near 0 or 1 --- in particular, the drift-audit rate (point estimate 0.67\%, $n{=}300$) where Wald can give a misleadingly symmetric interval or even negative lower bounds --- we use the Wilson score interval,
\[
\frac{\hat{p} + \frac{z^2}{2n} \;\pm\; z\sqrt{\frac{\hat{p}(1-\hat{p})}{n} + \frac{z^2}{4n^2}}}{1 + \frac{z^2}{n}},
\]
with $z{=}1.96$. Wilson is asymmetric near the boundaries, always lies in $[0, 1]$, and is the recommended interval for low-rate proportions.

The intervals capture sampling variability of $\hat{p}$ over the fixed test set; they do not capture variability over models, training runs, or seeds.

\paragraph{Scene-clustered bootstrap.} Wald and Wilson treat instances as independent, but CompoSET instances cluster within 80 base scenes and outcomes are correlated within scene. To account for this, we resample the 80 scenes with replacement ($B{=}10{,}000$, seed 42) and recompute the score as the pooled mean over all rows of the drawn scenes; a scene drawn $k$ times contributes its rows $k$-fold. We report percentile intervals over the bootstrap distribution. Clustering matters: the resulting intervals are 27--71\% wider than the row-level Wald intervals of Table~\ref{tab:app-full-results}. The conclusion is unchanged --- every one of the 22 dual-encoder Group upper bounds remains below the 16.67\% chance level (maximum 15.55\%, SigLIP-2 SO400M/14). The script is archived at \texttt{paper/analysis/scene\_edit\_hard/bootstrap\_scene\_clustered.py}.

\paragraph{Generator drift audit.} To bound how often the image-generation prompt (App.~\ref{app:imagegen}) leaks unintended changes into the foil, we sampled $N{=}300$ short-tier (base, foil) pairs and asked Gemini 2.5 Flash to flag any change beyond the labeled compositional edit. We manually verified every flag. Only 2/300 instances retained genuine multi-edit drift after verification: 0.67\%, Wilson 95\% CI $[0.18\%, 2.39\%]$. The remaining flagged cases were either disocclusion (newly visible background where an object was removed) or below-threshold lighting and blur differences. Gemini's flags concentrate in spatial edits (6 of 16 \texttt{modify\_spatial} and \texttt{swap\_spatial} pairs, against 13 of the other 284), where the generator must re-render the moved object's background.

\paragraph{Configs, seeds, hardware.} Evaluation code is at \url{https://github.com/laure-delisle/CompoSET}. The dataset is at \texttt{huggingface.co/datasets/CompoSET/CompoSET}. The full sweep is reproducible with \texttt{python -m eval.run\_all\_benchmarks -{}-models <list> -{}-benchmarks <list>}; per-model HF checkpoint paths and OpenCLIP \texttt{(architecture, pretrained-tag)} pairs are listed in App.~\ref{app:licenses}. CompoSET freeze hash: \texttt{662230d}. Dual-encoder evaluation is deterministic per checkpoint (no sampling); generative open-weight scoring uses caption log-likelihood with greedy decoding (no seed dependence); frontier prompted-pairwise selection uses temperature 0 with two API repeats per instance for counterbalancing. Dataset-generation seeds are not classical numerical seeds: the captioning and image-generation models are accessed via API (Anthropic Claude, Google Gemini), so reruns are not bit-identical. Hardware: a single NVIDIA RTX 4090 (24~GB) for all dual-encoder and open-weight generative evaluations. Productive wall-clock for the model sweep was $\sim$8--9 GPU-hours (dual-encoder cosine sweep $\sim$6--7 hr, open-weight generative slice $\sim$2 hr). End-to-end including debugging, schema migrations, and OOM patches over $\sim$5 days totalled $\sim$11--12 GPU-hours. Frontier scoring via API incurred no local compute and $<\$10$ in API spend.

\section{Licenses for Existing Assets}
\label{app:licenses}

Table~\ref{tab:licenses-models} lists the model checkpoints we evaluate and Table~\ref{tab:licenses-benchmarks} lists the comparison benchmarks. We use every asset under the license posted on its Hugging Face card (or the asset's own repository) and cite the original publication.

We load all dual-encoder checkpoints via OpenCLIP, which is itself MIT-licensed but operates only as a loader; each checkpoint is governed by the license of its upstream release. Models accessed outside OpenCLIP carry their own licenses, and frontier API models are governed by their providers' terms of service. The Source column reports OpenCLIP's \texttt{(architecture / pretrained-tag)} pair where applicable, which resolves to the corresponding upstream checkpoint at load time.

\begin{table}[H]
\centering
\caption{Licenses for evaluated model checkpoints. Frontier API models are governed by their providers' terms of service.}
\label{tab:licenses-models}
\footnotesize
\setlength{\tabcolsep}{4pt}
\resizebox{\textwidth}{!}{%
\begin{tabular}{@{}lllc@{}}
\toprule
Model & Source & License & Citation \\
\midrule
\multicolumn{4}{l}{\emph{Dual-encoder, contrastive baselines}} \\
CLIP B/32                                 & \texttt{openai/clip-vit-base-patch32}          & MIT          & \cite{radford2021clip} \\
CLIP B/16 (OpenAI)                        & \texttt{openai/clip-vit-base-patch16}          & MIT          & \cite{radford2021clip} \\
CLIP L/14 (OpenAI)                        & \texttt{openai/clip-vit-large-patch14}         & MIT          & \cite{radford2021clip} \\
CLIP B/16 LAION-400M                      & OpenCLIP \texttt{laion400m\_e32}               & MIT          & \cite{cherti2023reproducible} \\
CLIP B/16 LAION-2B                        & OpenCLIP \texttt{laion2b\_s34b\_b88k}          & MIT          & \cite{cherti2023reproducible} \\
CLIP L/14 LAION-2B                        & OpenCLIP \texttt{laion2b\_s32b\_b82k}          & MIT          & \cite{cherti2023reproducible} \\
CLIP H/14 LAION-2B                        & OpenCLIP \texttt{laion2b\_s32b\_b79k}          & MIT          & \cite{cherti2023reproducible} \\
SigLIP B/16                               & OpenCLIP \texttt{ViT-B-16-SigLIP / webli}      & Apache-2.0   & \cite{zhai2023siglip} \\
SigLIP L/16-384                           & OpenCLIP \texttt{ViT-L-16-SigLIP-384 / webli}  & Apache-2.0   & \cite{zhai2023siglip} \\
SigLIP SO400M/14-384                      & OpenCLIP \texttt{ViT-SO400M-14-SigLIP-384 / webli} & Apache-2.0   & \cite{zhai2023siglip} \\
SigLIP-2 B/16                             & OpenCLIP \texttt{ViT-B-16-SigLIP2 / webli}     & Apache-2.0   & \cite{tschannen2025siglip2} \\
SigLIP-2 L/16-384                         & OpenCLIP \texttt{ViT-L-16-SigLIP2-384 / webli} & Apache-2.0   & \cite{tschannen2025siglip2} \\
SigLIP-2 SO400M/14-378                    & OpenCLIP \texttt{ViT-SO400M-14-SigLIP2-378 / webli} & Apache-2.0   & \cite{tschannen2025siglip2} \\
\midrule
\multicolumn{4}{l}{\emph{Hard-negative compositional training}} \\
NegCLIP B/32                              & \texttt{Nano1337/negclip} (HF mirror)          & MIT          & \cite{yuksekgonul2023bow} \\
CLoVe B/32                                & \texttt{github.com/Netflix/clove}              & MIT          & \cite{castro2024clove} \\
\midrule
\multicolumn{4}{l}{\emph{Caption-enrichment pretraining}} \\
DreamLIP B/16 (30M)                       & \texttt{github.com/zyf0619sjtu/DreamLIP}       & CC-BY-4.0    & \cite{zheng2024dreamlip} \\
\midrule
\multicolumn{4}{l}{\emph{Data-curation recipes}} \\
DFN B/16                                  & OpenCLIP \texttt{dfn2b}                        & \texttt{apple-amlr} & \cite{fang2024dfn} \\
DFN H/14                                  & OpenCLIP \texttt{dfn5b}                        & \texttt{apple-amlr} & \cite{fang2024dfn} \\
EVA-02 B/16                               & OpenCLIP \texttt{merged2b\_s8b\_b131k}         & MIT          & \cite{fang2024eva02} \\
EVA-02 L/14                               & OpenCLIP \texttt{merged2b\_s4b\_b131k}         & MIT          & \cite{fang2024eva02} \\
MetaCLIP-2 H/14                           & OpenCLIP \texttt{metaclip2\_worldwide}         & CC-BY-NC 4.0 & \cite{chuang2025metaclip2} \\
Jina Embeddings v4                        & \texttt{jinaai/jina-embeddings-v4}             & Qwen Research License & \cite{jina2025embeddingsv4} \\
\midrule
\multicolumn{4}{l}{\emph{Generative open-weight (caveated slice)}} \\
BLIP-2 2.7B                               & \texttt{Salesforce/blip2-opt-2.7b}             & MIT          & \cite{li2023blip2} \\
LLaVA-1.6 Mistral 7B                      & \texttt{liuhaotian/llava-v1.6-mistral-7b}      & Apache-2.0   & \cite{liu2024llavanext} \\
Qwen2-VL 7B                               & \texttt{Qwen/Qwen2-VL-7B-Instruct}             & Apache-2.0   & \cite{wang2024qwen2vl} \\
\midrule
\multicolumn{4}{l}{\emph{Frontier (API)}} \\
Claude Sonnet 4.6                         & Anthropic API                                  & Anthropic ToS & \cite{anthropic2025claude} \\
Claude Haiku 4.5                          & Anthropic API                                  & Anthropic ToS & \cite{claudehaiku45} \\
Gemini 2.5 Flash                          & Google API                                     & Google ToS    & \cite{gemini25flash} \\
\bottomrule
\end{tabular}%
}
\end{table}

\begin{table}[H]
\centering
\caption{Licenses for comparison compositionality benchmarks.}
\label{tab:licenses-benchmarks}
\footnotesize
\setlength{\tabcolsep}{4pt}
\begin{tabular}{@{}lllc@{}}
\toprule
Benchmark & Source & License & Citation \\
\midrule
Winoground         & \texttt{facebook/winoground}            & Research use (HF gated)          & \cite{thrush2022winoground} \\
BiVLC              & \texttt{imirandam/BiVLC}                & MIT                              & \cite{miranda2024bivlc} \\
ColorSwap          & \texttt{stanfordnlp/colorswap}          & Apache-2.0                       & \cite{burapacheep2024colorswap} \\
NaturalBench       & \texttt{BaiqiL/NaturalBench}            & CC-BY-4.0                        & \cite{li2024naturalbench} \\
MMVP               & \texttt{MMVP/MMVP}                      & MIT                              & \cite{tong2024mmvp} \\
ColorBench         & \texttt{Liang-Yike/colorbench-robust}   & MIT                              & \cite{liang2025colorbench} \\
SugarCrepe         & \texttt{HuggingFaceM4/SugarCrepe}       & MIT                              & \cite{hsieh2023sugarcrepe} \\
SugarCrepe++       & \texttt{aburns4/SugarCrepe\_pp}         & MIT                              & \cite{dumpala2024sugarcrepepp} \\
What's Up          & \texttt{ServiceNow/whatsup\_all}        & MIT                              & \cite{kamath2023whatsup} \\
\bottomrule
\end{tabular}
\end{table}

\section{What's Up: Preposition-Level Bias in Dual-Encoders}
\label{app:whatsup-per-prep}

The What's Up Group score in Table~\ref{tab:main-results} is well below chance for every dual-encoder we evaluated. This appendix breaks the score down per preposition for CLIP-B/32 and shows that the cause is a fixed preposition bias: the model assigns one preposition higher similarity than its opposite regardless of the image, so single-direction (1-of-2) accuracy averages to roughly chance while strict bidirectional Group collapses far below chance. Table~\ref{tab:whatsup-per-prep} reports per-preposition I2T and T2I accuracy alongside the bidirectional strict-AND scores aggregated per axis.

\begin{table}[H]
\centering
\caption{CLIP-B/32 on WhatsUp, broken down by preposition. \emph{I2T} and \emph{T2I} are 1-of-2 accuracy where the foil is the opposite preposition on the same axis (chance 50\%); \emph{Group} is full bidirectional Winoground Group (strict-AND of I2T and T2I; chance 16.7\%) over the (prep$_a$, prep$_b$) pair. CLIP-B/32 ranks one preposition above its opposite regardless of image content (e.g., \emph{on}: 92.2\% I2T vs.\ \emph{under}: 22.3\%), so per-axis Group collapses far below chance.}
\label{tab:whatsup-per-prep}
\small
\begin{tabular}{llrrrrr}
\toprule
 & & \multicolumn{2}{c}{Per direction} & \multicolumn{3}{c}{Bidirectional} \\
\cmidrule(lr){3-4}\cmidrule(lr){5-7}
Pair & Preposition & I2T & T2I & I2T & T2I & Group \\
\midrule
\multirow{2}{*}{on / under} & on & 92.2 & 74.8 & \multirow{2}{*}{14.6} & \multirow{2}{*}{13.6} & \multirow{2}{*}{3.9} \\
 & under & 22.3 & 36.9 & & & \\
\midrule
\multirow{2}{*}{left of / right of} & left of & 90.7 & 49.8 & \multirow{2}{*}{4.9} & \multirow{2}{*}{2.9} & \multirow{2}{*}{0.0} \\
 & right of & 11.2 & 51.7 & & & \\
\midrule
\multirow{2}{*}{in front of / behind} & in front of & 80.4 & 38.2 & \multirow{2}{*}{3.9} & \multirow{2}{*}{2.9} & \multirow{2}{*}{1.0} \\
 & behind & 10.8 & 63.7 & & & \\
\midrule
\multicolumn{2}{l}{Overall} & 51.2 & 52.1 & 7.1 & 5.6 & 1.2 \\
\bottomrule
\end{tabular}
\end{table}

\section{Dataset Property Analyses}
\label{app:dataset-properties}

This appendix reports three properties of CompoSET as a \emph{dataset}, characterized independently of any specific model: how visually different the paired images are (\S\ref{app:image-deviation}), how textually different the paired captions are across verbosity tiers (\S\ref{app:text-deviation}), and how visually rich the underlying scenes are (\S\ref{app:objcount}). The within-vs-across embedding-space property is summarized in §\ref{sec:stats-cosine} and Figure~\ref{fig:analysis-trio}, and not duplicated here. Each appendix supports a claim made in §\ref{sec:analysis} of the main paper.

\subsection{Image-Deviation \& Pixel-Affected Analysis}
\label{app:image-deviation}

\begin{figure}[H]
\centering
\includegraphics[width=\textwidth]{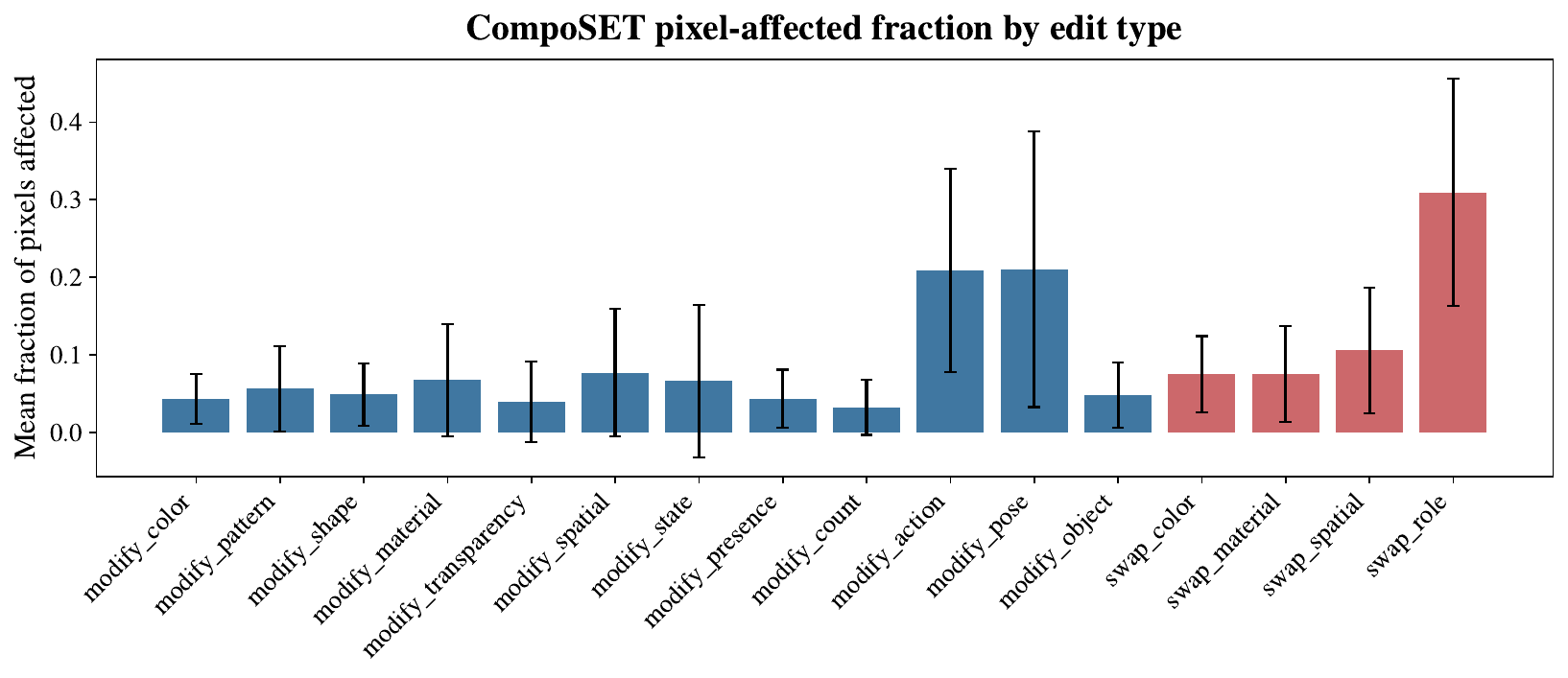}
\caption{Per-edit-type pixel-affected fraction within CompoSET. Identity edits (color, pattern, material) are the most localized; pose, action, and role edits perturb larger image regions.}
\label{fig:imagedev-per-edit}
\end{figure}

For every base/foil pair in CompoSET (1{,}776 pairs) and a seed-42 200-pair subsample of each prior benchmark, we resize images to $512\times512$ and compute pixel-level (MAE, $L_2$, $L_\infty$, fraction of pixels with channel-max difference above $10/255$ after $\sigma{=}1.0$ Gaussian blur), structural (SSIM, MS-SSIM), perceptual (LPIPS-AlexNet, LPIPS-VGG, DISTS), and semantic (DINOv2-base CLS cosine, CLIP ViT-B/32 image cosine) metrics. The headline ECDF is in Figure~\ref{fig:analysis-trio}, left. The per-edit-type breakdown (Figure~\ref{fig:imagedev-per-edit}) is consistent with construction intent: \texttt{modify\_color} and \texttt{modify\_pattern} have the smallest fraction-affected ($\sim$3--5\%), while \texttt{modify\_spatial} and \texttt{modify\_count} are largest within CompoSET ($\sim$15\%) but still well below the prior-benchmark scale.

\subsection{Text-Deviation Analysis}
\label{app:text-deviation}

\begin{figure}[H]
\centering
\includegraphics[width=\textwidth]{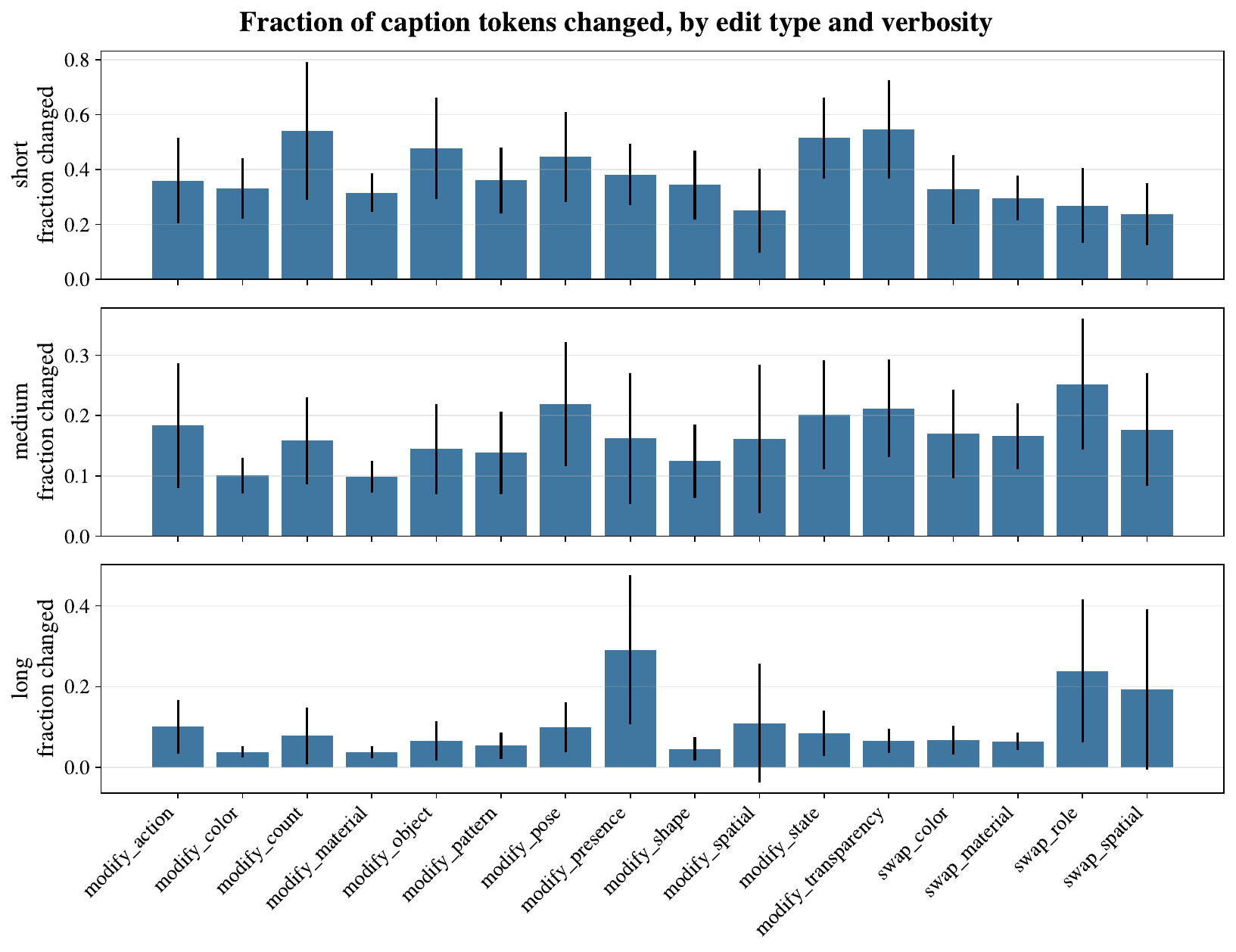}
\caption{Fraction of caption tokens changed in CompoSET by edit type and verbosity tier. Action and state edits are wider text changes; color and pattern edits are single-token swaps.}
\label{fig:textdev-per-edit}
\end{figure}

For every CompoSET caption pair at each of the three verbosity tiers (short, medium, long) and a per-benchmark sample of prior benchmarks (Winoground, BiVLC, What's~Up, SugarCrepe, MMVP), we compute token-level Levenshtein distance using CLIP ViT-B/32 BPE tokens, fraction of tokens changed, and sentence-embedding cosine using \texttt{all-mpnet-base-v2}. The headline ECDF for the medium tier is in Figure~\ref{fig:analysis-trio}, middle. Across tiers, mean fraction of tokens changed: \emph{short} 0.38, \emph{medium} 0.16, \emph{long} 0.09; MPNet cosine: 0.85, 0.92, 0.97. The per-edit-type breakdown (Figure~\ref{fig:textdev-per-edit}) shows \texttt{modify\_action} and \texttt{modify\_state} as the largest text changes within tier (multi-word predicate edits), while \texttt{modify\_color} and \texttt{modify\_pattern} are the smallest (single-token attribute swaps).

\subsection{Object-Count Comparison}
\label{app:objcount}

\textbf{Setup.} We run OWLv2-base (\texttt{google/owlv2-base-patch16-ensemble}) with score threshold 0.3 and class-agnostic NMS at IoU 0.5, prompted with a 116-class open-vocabulary list (everyday objects, furniture, vehicles, animals, food, etc.). We apply the detector identically to CompoSET (50-scene subsample), Winoground, BiVLC, and SugarCrepe (200-image samples each, seed 42). We count detected instances after class-agnostic NMS, so repeated objects are rewarded (three chairs counts as three). We report two summaries: excluding people (primary, since person-density varies across benchmarks) and including people.

CompoSET scenes contain 20.7 detected objects per image on average (median 16, IQR 11.2--26.8, max 66). Winoground averages 5.1, BiVLC 6.1, SugarCrepe 6.3. CompoSET's 25\textsuperscript{th} percentile (11.2 objects per image) exceeds the 75\textsuperscript{th} percentile of every comparison benchmark (Table~\ref{tab:objcount-distrib}). Including people, the gap holds: CompoSET 23.4, Winoground 5.9, BiVLC 6.8, SugarCrepe 7.2.

\begin{table}[H]
\centering
\caption{\textbf{CompoSET is $3$--$4\times$ more multi-object than every comparison benchmark, both excluding and including people.} Per-image detected-object counts under an identical OWLv2 protocol.}
\label{tab:objcount-distrib}
\small
\begin{tabular}{lrrrrrr}
\toprule
Benchmark & mean & median & std & p25 & p75 & max \\
\midrule
\multicolumn{7}{l}{\emph{Excluding people (primary)}} \\
\textbf{CompoSET} & \textbf{20.7} & \textbf{16.0} & 13.5 & \textbf{11.2} & \textbf{26.8} & 66 \\
Winoground        & 5.1           & 3.0           & 8.5  & 1.0           & 6.0           & 76 \\
BiVLC             & 6.1           & 4.0           & 7.2  & 2.0           & 7.0           & 63 \\
SugarCrepe        & 6.3           & 4.0           & 6.4  & 2.0           & 8.0           & 46 \\
\midrule
\multicolumn{7}{l}{\emph{Including people}} \\
\textbf{CompoSET} & \textbf{23.4} & \textbf{19.0} & 14.3 & \textbf{13.0} & \textbf{30.8} & 69 \\
Winoground        & 5.9           & 3.0           & 8.7  & 2.0           & 6.0           & 77 \\
BiVLC             & 6.8           & 5.0           & 7.3  & 3.0           & 8.0           & 63 \\
SugarCrepe        & 7.2           & 5.0           & 7.1  & 3.0           & 8.0           & 48 \\
\bottomrule
\end{tabular}
\end{table}

\textbf{Detector caveats.} The claim is relative, not absolute: under an identical detection protocol, CompoSET images contain more detected instances than Winoground / BiVLC / SugarCrepe. We anticipate four pushbacks. \emph{Detector bias}: OWLv2 may systematically under- or over-detect on any given benchmark, but using the same model, prompt list, and threshold across benchmarks makes comparisons relative rather than absolute. \emph{Threshold sensitivity}: re-evaluating the stored raw detections at thresholds 0.2 and 0.4 preserves the benchmark ordering. \emph{Vocabulary coverage}: the 116-class prompt list misses exotic objects, which under-counts CompoSET more than the comparison benchmarks because CompoSET scenes are deliberately diverse; this biases the comparison \emph{against} our claim, not in favor of it. \emph{Sample size}: at $n{=}200$ per comparator (seed 42), the standard error on the mean count is well below the effect size; Winoground's full size is 800 images, so the comparator sample is 25\% of that benchmark.

\section{Model Diagnostics}
\label{app:model-diagnostics}

This appendix reports four analyses of \emph{model behavior} on CompoSET: the full 22-dual-encoder pool with confidence intervals across all 10 benchmarks (\S\ref{app:full-results}), per-edit-type and per-verbosity performance breakdowns on CompoSET (\S\ref{app:per-edittype-per-verbosity}), per-pair cosine-margin distributions exposing where dual encoders confidently fail (\S\ref{app:cosine-margin}), and a separately scored slice of generative VLMs (open-weight and frontier API; \S\ref{app:gen-vlm}).

\subsection{Full 22-Dual-Encoder Results}
\label{app:full-results}

Table~\ref{tab:app-full-results} reports the full 22-dual-encoder pool across all 10 compositionality benchmarks, with 95\% Wald confidence intervals (\S\ref{app:repro}). The 14 models highlighted in Table~\ref{tab:main-results} are a representative subset; the remaining 8 (additional CLIP variants, SigLIP L/16-384, DFN B/16, EVA-02 B/16, Jina-v4) follow the same overall pattern: every dual encoder is at or below chance on CompoSET, while most are above chance on prior benchmarks other than Winoground and What's Up.

The direction split is one-sided. Point estimates of T2I exceed I2T for every one of the 22 dual encoders. I2T is at or below the 25\% direction chance for all 22; the best, SigLIP-2 SO400M/14, reaches 24.5\% (scene-clustered 95\% CI $[22.5, 26.5]$, \S\ref{app:repro}), indistinguishable from chance. T2I rises modestly above chance only at the top of the pool: the SigLIP family spans 25.6--29.6\% (all members except SigLIP B/16, at 24.3\%), led by SigLIP-2 SO400M/14 at 29.6\% (clustered CI $[27.5, 31.7]$, above chance); DreamLIP B/16 is the only other encoder above 25\% (26.4\%). The generative slice (\S\ref{app:gen-vlm}) shows the opposite asymmetry: under caption log-likelihood scoring, I2T far exceeds T2I. The two scoring primitives fail differently, and neither direction alone carries the Group failure.

\begin{table}[H]
\centering
\caption{Full 22-dual-encoder pool across 10 compositionality benchmarks. Each cell reports point estimate (\%) with 95\% Wald confidence interval $[\,$lo, hi$\,]$ (\S\ref{app:repro}). Best-in-pool per column in \textbf{bold}. Metric per benchmark: Group for Winoground-style 2$\times$2 (CompoSET, Winoground, BiVLC, What's Up, ColorSwap, MMVP); T2I for image-foil-only benchmarks (NaturalBench, ColorBench); I2T for text-foil-only benchmarks (SugarCrepe, SugarCrepe$^{++}$).}
\label{tab:app-full-results}
\scriptsize
\setlength{\tabcolsep}{3pt}
\resizebox{\textwidth}{!}{%
\begin{tabular}{l>{\columncolor{black!7}\centering\arraybackslash}p{36pt}>{\centering\arraybackslash}p{36pt}>{\centering\arraybackslash}p{36pt}>{\centering\arraybackslash}p{36pt}>{\centering\arraybackslash}p{36pt}>{\centering\arraybackslash}p{36pt}>{\centering\arraybackslash}p{36pt}>{\centering\arraybackslash}p{36pt}>{\centering\arraybackslash}p{36pt}>{\centering\arraybackslash}p{36pt}}
\toprule
Model & \makecell{\textbf{Compo}\\\textbf{SET}} & \makecell{Wino\\ground} & BiVLC & \makecell{What's\\Up} & \makecell{Color\\Swap} & MMVP & \makecell{Natural\\Bench} & \makecell{Color\\Bench} & \makecell{Sugar\\Crepe} & \makecell{Sugar\\Crepe\textsuperscript{++}} \\
\midrule
\emph{Chance} & \emph{16.7} & \emph{16.7} & \emph{16.7} & \emph{16.7} & \emph{16.7} & \emph{16.7} & \emph{50.0} & \emph{50.0} & \emph{50.0} & \emph{50.0} \\
\midrule
\multicolumn{11}{l}{\emph{Contrastive baselines}} \\
CLIP B/32 & \makecell{4.4\\\scriptsize[3.9,5.0]} & \makecell{7.5\\\scriptsize[5.3,10.5]} & \makecell{47.2\\\scriptsize[45.4,49.0]} & \makecell{2.2\\\scriptsize[1.5,3.2]} & \makecell{8.7\\\scriptsize[6.0,12.4]} & \makecell{8.1\\\scriptsize[4.6,14.0]} & \makecell{68.1\\\scriptsize[66.4,69.8]} & \makecell{46.7\\\scriptsize[45.0,48.3]} & \makecell{78.6\\\scriptsize[77.4,79.7]} & \makecell{65.5\\\scriptsize[64.1,66.8]} \\
CLIP B/16 (OpenAI) & \makecell{4.8\\\scriptsize[4.2,5.4]} & \makecell{6.8\\\scriptsize[4.7,9.6]} & \makecell{47.8\\\scriptsize[46.0,49.6]} & \makecell{3.0\\\scriptsize[2.2,4.1]} & \makecell{7.7\\\scriptsize[5.2,11.2]} & \makecell{10.4\\\scriptsize[6.3,16.7]} & \makecell{69.1\\\scriptsize[67.4,70.8]} & \makecell{48.8\\\scriptsize[47.1,50.4]} & \makecell{78.5\\\scriptsize[77.4,79.7]} & \makecell{65.1\\\scriptsize[63.7,66.4]} \\
CLIP B/16 (LAION-400M) & \makecell{8.4\\\scriptsize[7.6,9.1]} & \makecell{7.8\\\scriptsize[5.5,10.8]} & \makecell{54.2\\\scriptsize[52.4,56.0]} & \makecell{1.2\\\scriptsize[0.7,2.0]} & \makecell{15.3\\\scriptsize[11.7,19.8]} & \makecell{12.6\\\scriptsize[8.0,19.2]} & \makecell{70.4\\\scriptsize[68.8,72.1]} & \makecell{56.2\\\scriptsize[54.6,57.9]} & \makecell{78.5\\\scriptsize[77.3,79.6]} & \makecell{66.6\\\scriptsize[65.2,67.9]} \\
CLIP B/16 (LAION-2B) & \makecell{7.8\\\scriptsize[7.1,8.5]} & \makecell{8.2\\\scriptsize[5.9,11.4]} & \makecell{53.3\\\scriptsize[51.4,55.1]} & \makecell{1.1\\\scriptsize[0.6,1.8]} & \makecell{19.3\\\scriptsize[15.3,24.2]} & \makecell{12.6\\\scriptsize[8.0,19.2]} & \makecell{70.7\\\scriptsize[69.0,72.3]} & \makecell{50.6\\\scriptsize[49.0,52.2]} & \makecell{81.3\\\scriptsize[80.2,82.4]} & \makecell{70.1\\\scriptsize[68.7,71.4]} \\
CLIP L/14 (OpenAI) & \makecell{5.0\\\scriptsize[4.5,5.7]} & \makecell{6.8\\\scriptsize[4.7,9.6]} & \makecell{51.4\\\scriptsize[49.6,53.2]} & \makecell{3.4\\\scriptsize[2.5,4.6]} & \makecell{6.0\\\scriptsize[3.8,9.3]} & \makecell{7.4\\\scriptsize[4.1,13.1]} & \makecell{70.1\\\scriptsize[68.4,71.8]} & \makecell{49.9\\\scriptsize[48.3,51.6]} & \makecell{79.5\\\scriptsize[78.3,80.6]} & \makecell{68.2\\\scriptsize[66.8,69.5]} \\
CLIP L/14 (LAION-2B) & \makecell{8.4\\\scriptsize[7.7,9.2]} & \makecell{8.8\\\scriptsize[6.4,11.9]} & \makecell{55.0\\\scriptsize[53.2,56.8]} & \makecell{2.1\\\scriptsize[1.4,3.1]} & \makecell{24.0\\\scriptsize[19.5,29.1]} & \makecell{18.5\\\scriptsize[12.9,25.9]} & \makecell{72.9\\\scriptsize[71.3,74.5]} & \makecell{51.5\\\scriptsize[49.9,53.1]} & \makecell{81.3\\\scriptsize[80.1,82.4]} & \makecell{71.5\\\scriptsize[70.2,72.7]} \\
CLIP H/14 (LAION-2B) & \makecell{8.4\\\scriptsize[7.7,9.1]} & \makecell{10.5\\\scriptsize[7.9,13.9]} & \makecell{56.9\\\scriptsize[55.1,58.7]} & \makecell{0.5\\\scriptsize[0.2,1.1]} & \makecell{32.0\\\scriptsize[27.0,37.5]} & \makecell{18.5\\\scriptsize[12.9,25.9]} & \makecell{73.5\\\scriptsize[71.9,75.1]} & \makecell{55.4\\\scriptsize[53.8,57.1]} & \makecell{82.5\\\scriptsize[81.4,83.5]} & \makecell{70.9\\\scriptsize[69.6,72.2]} \\
SigLIP B/16 & \makecell{10.2\\\scriptsize[9.4,11.0]} & \makecell{10.2\\\scriptsize[7.6,13.6]} & \makecell{59.6\\\scriptsize[57.8,61.4]} & \makecell{1.6\\\scriptsize[1.1,2.5]} & \makecell{30.7\\\scriptsize[25.7,36.1]} & \makecell{23.0\\\scriptsize[16.7,30.7]} & \makecell{74.7\\\scriptsize[73.1,76.2]} & \makecell{55.4\\\scriptsize[53.8,57.0]} & \makecell{81.3\\\scriptsize[80.2,82.4]} & \makecell{70.3\\\scriptsize[68.9,71.5]} \\
SigLIP L/16-384 & \makecell{12.7\\\scriptsize[11.9,13.7]} & \makecell{13.0\\\scriptsize[10.1,16.7]} & \makecell{\textbf{61.7}\\\scriptsize[59.9,63.4]} & \makecell{1.9\\\scriptsize[1.2,2.8]} & \makecell{42.3\\\scriptsize[36.9,48.0]} & \makecell{28.1\\\scriptsize[21.2,36.3]} & \makecell{77.4\\\scriptsize[75.8,78.9]} & \makecell{59.2\\\scriptsize[57.6,60.8]} & \makecell{83.2\\\scriptsize[82.1,84.2]} & \makecell{73.6\\\scriptsize[72.3,74.8]} \\
SigLIP SO400M/14 & \makecell{13.5\\\scriptsize[12.6,14.4]} & \makecell{12.2\\\scriptsize[9.4,15.8]} & \makecell{60.0\\\scriptsize[58.2,61.8]} & \makecell{2.8\\\scriptsize[2.1,3.9]} & \makecell{37.0\\\scriptsize[31.7,42.6]} & \makecell{28.9\\\scriptsize[21.9,37.0]} & \makecell{77.2\\\scriptsize[75.6,78.7]} & \makecell{56.9\\\scriptsize[55.3,58.5]} & \makecell{83.6\\\scriptsize[82.5,84.6]} & \makecell{73.7\\\scriptsize[72.5,75.0]} \\
SigLIP-2 B/16 & \makecell{11.3\\\scriptsize[10.5,12.2]} & \makecell{10.8\\\scriptsize[8.1,14.2]} & \makecell{57.6\\\scriptsize[55.8,59.3]} & \makecell{2.1\\\scriptsize[1.4,3.1]} & \makecell{34.7\\\scriptsize[29.5,40.2]} & \makecell{23.7\\\scriptsize[17.3,31.5]} & \makecell{75.9\\\scriptsize[74.3,77.4]} & \makecell{56.0\\\scriptsize[54.3,57.6]} & \makecell{84.1\\\scriptsize[83.0,85.1]} & \makecell{75.4\\\scriptsize[74.2,76.6]} \\
SigLIP-2 L/16-384 & \makecell{13.6\\\scriptsize[12.7,14.6]} & \makecell{14.0\\\scriptsize[10.9,17.7]} & \makecell{52.8\\\scriptsize[51.0,54.6]} & \makecell{2.3\\\scriptsize[1.6,3.3]} & \makecell{44.3\\\scriptsize[38.8,50.0]} & \makecell{23.0\\\scriptsize[16.7,30.7]} & \makecell{78.8\\\scriptsize[77.2,80.2]} & \makecell{\textbf{61.4}\\\scriptsize[59.9,63.0]} & \makecell{83.3\\\scriptsize[82.2,84.3]} & \makecell{74.6\\\scriptsize[73.3,75.8]} \\
SigLIP-2 SO400M/14 & \makecell{\textbf{13.9}\\\scriptsize[13.0,14.9]} & \makecell{\textbf{15.2}\\\scriptsize[12.1,19.1]} & \makecell{49.5\\\scriptsize[47.7,51.3]} & \makecell{4.2\\\scriptsize[3.2,5.5]} & \makecell{39.7\\\scriptsize[34.3,45.3]} & \makecell{\textbf{31.1}\\\scriptsize[23.9,39.4]} & \makecell{\textbf{79.4}\\\scriptsize[77.9,80.9]} & \makecell{54.3\\\scriptsize[52.7,55.9]} & \makecell{84.3\\\scriptsize[83.2,85.3]} & \makecell{\textbf{75.8}\\\scriptsize[74.5,77.0]} \\
\midrule
\multicolumn{11}{l}{\emph{Hard-negative compositional training}} \\
NegCLIP B/32 & \makecell{7.0\\\scriptsize[6.4,7.7]} & \makecell{8.0\\\scriptsize[5.7,11.1]} & \makecell{55.3\\\scriptsize[53.5,57.1]} & \makecell{4.1\\\scriptsize[3.1,5.3]} & \makecell{14.0\\\scriptsize[10.5,18.4]} & \makecell{12.6\\\scriptsize[8.0,19.2]} & \makecell{72.1\\\scriptsize[70.5,73.7]} & \makecell{43.0\\\scriptsize[41.4,44.6]} & \makecell{83.4\\\scriptsize[82.3,84.4]} & \makecell{67.4\\\scriptsize[66.0,68.7]} \\
CLoVe B/32 & \makecell{5.5\\\scriptsize[5.0,6.2]} & \makecell{4.8\\\scriptsize[3.1,7.3]} & \makecell{48.3\\\scriptsize[46.5,50.1]} & \makecell{0.8\\\scriptsize[0.4,1.5]} & \makecell{15.0\\\scriptsize[11.4,19.5]} & \makecell{11.1\\\scriptsize[6.8,17.5]} & \makecell{67.8\\\scriptsize[66.1,69.5]} & \makecell{45.8\\\scriptsize[44.2,47.5]} & \makecell{\textbf{84.9}\\\scriptsize[83.8,85.9]} & \makecell{56.1\\\scriptsize[54.7,57.6]} \\
\midrule
\multicolumn{11}{l}{\emph{Caption-enrichment pretraining}} \\
DreamLIP B/16 (30M) & \makecell{13.7\\\scriptsize[12.8,14.6]} & \makecell{11.5\\\scriptsize[8.7,15.0]} & \makecell{53.9\\\scriptsize[52.1,55.7]} & \makecell{\textbf{5.4}\\\scriptsize[4.3,6.9]} & \makecell{\textbf{61.0}\\\scriptsize[55.4,66.3]} & \makecell{16.3\\\scriptsize[11.0,23.4]} & \makecell{74.4\\\scriptsize[72.8,75.9]} & \makecell{48.6\\\scriptsize[46.9,50.2]} & \makecell{81.5\\\scriptsize[80.4,82.6]} & \makecell{66.2\\\scriptsize[64.9,67.6]} \\
\midrule
\multicolumn{11}{l}{\emph{Data-curation recipes}} \\
DFN B/16 & \makecell{8.4\\\scriptsize[7.7,9.2]} & \makecell{8.5\\\scriptsize[6.1,11.6]} & \makecell{55.1\\\scriptsize[53.3,56.9]} & \makecell{1.1\\\scriptsize[0.6,1.8]} & \makecell{22.0\\\scriptsize[17.7,27.0]} & \makecell{21.5\\\scriptsize[15.4,29.1]} & \makecell{72.3\\\scriptsize[70.7,73.9]} & \makecell{51.7\\\scriptsize[50.1,53.3]} & \makecell{79.4\\\scriptsize[78.2,80.5]} & \makecell{69.0\\\scriptsize[67.7,70.3]} \\
DFN H/14 & \makecell{10.8\\\scriptsize[10.0,11.7]} & \makecell{11.0\\\scriptsize[8.3,14.4]} & \makecell{59.4\\\scriptsize[57.6,61.1]} & \makecell{3.5\\\scriptsize[2.6,4.7]} & \makecell{45.0\\\scriptsize[39.5,50.7]} & \makecell{30.4\\\scriptsize[23.2,38.6]} & \makecell{75.8\\\scriptsize[74.2,77.3]} & \makecell{57.5\\\scriptsize[55.9,59.1]} & \makecell{84.0\\\scriptsize[83.0,85.0]} & \makecell{74.6\\\scriptsize[73.3,75.8]} \\
EVA-02 B/16 & \makecell{7.8\\\scriptsize[7.1,8.5]} & \makecell{9.0\\\scriptsize[6.6,12.2]} & \makecell{52.1\\\scriptsize[50.3,53.9]} & \makecell{0.7\\\scriptsize[0.4,1.4]} & \makecell{11.0\\\scriptsize[7.9,15.0]} & \makecell{16.3\\\scriptsize[11.0,23.4]} & \makecell{71.2\\\scriptsize[69.6,72.9]} & \makecell{53.4\\\scriptsize[51.8,55.0]} & \makecell{78.4\\\scriptsize[77.2,79.6]} & \makecell{68.6\\\scriptsize[67.2,69.9]} \\
EVA-02 L/14 & \makecell{8.6\\\scriptsize[7.9,9.4]} & \makecell{10.2\\\scriptsize[7.6,13.6]} & \makecell{55.2\\\scriptsize[53.4,57.0]} & \makecell{2.0\\\scriptsize[1.3,2.9]} & \makecell{12.7\\\scriptsize[9.4,16.9]} & \makecell{23.0\\\scriptsize[16.7,30.7]} & \makecell{73.3\\\scriptsize[71.6,74.9]} & \makecell{53.4\\\scriptsize[51.8,55.1]} & \makecell{80.9\\\scriptsize[79.7,81.9]} & \makecell{70.8\\\scriptsize[69.5,72.1]} \\
MetaCLIP-2 H/14 & \makecell{8.9\\\scriptsize[8.2,9.7]} & \makecell{11.0\\\scriptsize[8.3,14.4]} & \makecell{48.3\\\scriptsize[46.5,50.1]} & \makecell{1.6\\\scriptsize[1.1,2.5]} & \makecell{20.7\\\scriptsize[16.5,25.6]} & \makecell{17.0\\\scriptsize[11.6,24.3]} & \makecell{72.9\\\scriptsize[71.2,74.4]} & \makecell{53.2\\\scriptsize[51.6,54.8]} & \makecell{80.2\\\scriptsize[79.0,81.3]} & \makecell{69.3\\\scriptsize[68.0,70.6]} \\
Jina-v4 & \makecell{8.5\\\scriptsize[7.8,9.3]} & \makecell{5.8\\\scriptsize[3.9,8.5]} & \makecell{49.1\\\scriptsize[47.3,50.9]} & \makecell{2.1\\\scriptsize[1.4,3.1]} & \makecell{13.0\\\scriptsize[9.7,17.3]} & \makecell{8.1\\\scriptsize[4.6,14.0]} & \makecell{71.6\\\scriptsize[69.9,73.2]} & \makecell{36.3\\\scriptsize[34.7,37.8]} & \makecell{80.6\\\scriptsize[79.5,81.7]} & \makecell{71.2\\\scriptsize[69.9,72.4]} \\
\bottomrule
\end{tabular}%
}
\end{table}

\subsection{Per-Edit-Type \texorpdfstring{\ding{53}}{x} Per-Verbosity Breakdown}
\label{app:per-edittype-per-verbosity}

We report two views of CompoSET performance across the 16 edit types. Table~\ref{tab:app-per-edittype} gives Group score per (model, edit type), pooled across short/medium/long captions, with 95\% Wald confidence intervals (\S\ref{app:repro}); per-edit $n$ at this pool ranges from 135 (\texttt{swap\_role}) to 618 (\texttt{modify\_color}). Table~\ref{tab:app-per-edittype-per-tier} gives the same Group score split by verbosity tier and averaged across the 14 models, exposing which edit types degrade most as captions grow longer.

Identity \& Appearance edits separate models the most: \texttt{modify\_color} spans 20--52\% Group, with DreamLIP B/16 leading by a wide margin. Spatial and presence edits are uniformly hard ($\le 5$\%) regardless of model. \texttt{swap\_role} ($n{=}135$ pooled, $n{=}45$ per tier) has the widest CIs in the table; we keep it in benchmark totals but flag this caveat in §\ref{sec:discussion}. The per-tier panel (Table~\ref{tab:app-per-edittype-per-tier}) shows the verbosity drop concentrating on \texttt{modify\_color} ($35.3 \to 30.5$), \texttt{modify\_pose} ($10.4 \to 5.3$), and \texttt{modify\_action} ($27.2 \to 23.4$); a few edits move in the opposite direction at low absolute level (\texttt{modify\_presence} $0.3 \to 2.5$).

\begin{landscape}
\begin{table}[H]
\centering
\caption{Per-edit-type Group score (\%) on CompoSET for the 14 dual encoders of Table~\ref{tab:main-results}, pooled across the three caption verbosity levels. Each cell reports point estimate with 95\% Wald confidence interval (\S\ref{app:repro}); $n$ is per-edit-type instance count, summed across short/medium/long tiers ($3\times$ the per-scene count). Note the wide intervals on \texttt{swap\_role} ($n{=}135$ pooled, $n{=}45$ per tier).}
\label{tab:app-per-edittype}
\footnotesize
\setlength{\tabcolsep}{4pt}
\renewcommand{\arraystretch}{1.15}
\resizebox{\linewidth}{!}{%
\begin{tabular}{l r >{\centering\arraybackslash}p{42pt} >{\centering\arraybackslash}p{42pt} >{\centering\arraybackslash}p{42pt} >{\centering\arraybackslash}p{42pt} >{\centering\arraybackslash}p{42pt} >{\centering\arraybackslash}p{42pt} >{\centering\arraybackslash}p{42pt} >{\centering\arraybackslash}p{42pt} >{\centering\arraybackslash}p{42pt} >{\centering\arraybackslash}p{42pt} >{\centering\arraybackslash}p{42pt} >{\centering\arraybackslash}p{42pt} >{\centering\arraybackslash}p{42pt} >{\centering\arraybackslash}p{42pt}}
\toprule
Edit type & $n$ & \makecell{CLIP\\B/32} & \makecell{CLIP B/16\\LAION-2B} & \makecell{CLIP H/14\\LAION-2B} & \makecell{SigLIP\\B/16} & \makecell{SigLIP\\SO400M} & \makecell{SigLIP-2\\B/16} & \makecell{SigLIP-2\\L/16} & \makecell{SigLIP-2\\SO400M} & \makecell{NegCLIP\\B/32} & \makecell{CLoVe\\B/32} & \makecell{DreamLIP\\B/16} & \makecell{DFN\\H/14} & \makecell{EVA-02\\L/14} & \makecell{MetaCLIP-2\\H/14} \\
\midrule
\multicolumn{16}{l}{\emph{Identity}} \\
modify\_color & 618 & \makecell{20.4\\\scriptsize[17.2,23.6]} & \makecell{29.9\\\scriptsize[26.3,33.5]} & \makecell{28.8\\\scriptsize[25.2,32.4]} & \makecell{37.2\\\scriptsize[33.4,41.0]} & \makecell{39.5\\\scriptsize[35.6,43.3]} & \makecell{39.2\\\scriptsize[35.3,43.0]} & \makecell{38.5\\\scriptsize[34.7,42.3]} & \makecell{36.2\\\scriptsize[32.5,40.0]} & \makecell{30.6\\\scriptsize[26.9,34.2]} & \makecell{27.7\\\scriptsize[24.1,31.2]} & \makecell{\textbf{52.4}\\\scriptsize[48.5,56.4]} & \makecell{35.1\\\scriptsize[31.3,38.9]} & \makecell{26.5\\\scriptsize[23.1,30.0]} & \makecell{27.2\\\scriptsize[23.7,30.7]} \\
modify\_pattern & 303 & \makecell{6.3\\\scriptsize[3.5,9.0]} & \makecell{14.5\\\scriptsize[10.6,18.5]} & \makecell{9.6\\\scriptsize[6.3,12.9]} & \makecell{20.1\\\scriptsize[15.6,24.6]} & \makecell{21.5\\\scriptsize[16.8,26.1]} & \makecell{17.8\\\scriptsize[13.5,22.1]} & \makecell{16.8\\\scriptsize[12.6,21.0]} & \makecell{23.8\\\scriptsize[19.0,28.6]} & \makecell{8.6\\\scriptsize[5.4,11.7]} & \makecell{8.9\\\scriptsize[5.7,12.1]} & \makecell{\textbf{29.4}\\\scriptsize[24.2,34.5]} & \makecell{9.9\\\scriptsize[6.5,13.3]} & \makecell{14.2\\\scriptsize[10.3,18.1]} & \makecell{11.9\\\scriptsize[8.2,15.5]} \\
modify\_material & 483 & \makecell{1.7\\\scriptsize[0.5,2.8]} & \makecell{6.8\\\scriptsize[4.6,9.1]} & \makecell{8.3\\\scriptsize[5.8,10.7]} & \makecell{11.2\\\scriptsize[8.4,14.0]} & \makecell{\textbf{16.4}\\\scriptsize[13.1,19.7]} & \makecell{10.8\\\scriptsize[8.0,13.5]} & \makecell{14.7\\\scriptsize[11.5,17.9]} & \makecell{13.7\\\scriptsize[10.6,16.7]} & \makecell{4.3\\\scriptsize[2.5,6.2]} & \makecell{3.3\\\scriptsize[1.7,4.9]} & \makecell{12.2\\\scriptsize[9.3,15.1]} & \makecell{12.6\\\scriptsize[9.7,15.6]} & \makecell{8.9\\\scriptsize[6.4,11.4]} & \makecell{8.1\\\scriptsize[5.6,10.5]} \\
modify\_shape & 309 & \makecell{2.6\\\scriptsize[0.8,4.4]} & \makecell{8.7\\\scriptsize[5.6,11.9]} & \makecell{10.7\\\scriptsize[7.2,14.1]} & \makecell{10.0\\\scriptsize[6.7,13.4]} & \makecell{12.3\\\scriptsize[8.6,16.0]} & \makecell{11.0\\\scriptsize[7.5,14.5]} & \makecell{\textbf{14.6}\\\scriptsize[10.6,18.5]} & \makecell{12.6\\\scriptsize[8.9,16.3]} & \makecell{6.8\\\scriptsize[4.0,9.6]} & \makecell{4.5\\\scriptsize[2.2,6.8]} & \makecell{\textbf{14.6}\\\scriptsize[10.6,18.5]} & \makecell{9.4\\\scriptsize[6.1,12.6]} & \makecell{7.4\\\scriptsize[4.5,10.4]} & \makecell{12.0\\\scriptsize[8.4,15.6]} \\
modify\_transparency & 300 & \makecell{0.7\\\scriptsize[0.0,1.6]} & \makecell{1.7\\\scriptsize[0.2,3.1]} & \makecell{0.7\\\scriptsize[0.0,1.6]} & \makecell{1.3\\\scriptsize[0.0,2.6]} & \makecell{1.7\\\scriptsize[0.2,3.1]} & \makecell{\textbf{2.3}\\\scriptsize[0.6,4.0]} & \makecell{\textbf{2.3}\\\scriptsize[0.6,4.0]} & \makecell{1.0\\\scriptsize[0.0,2.1]} & \makecell{1.0\\\scriptsize[0.0,2.1]} & \makecell{0.7\\\scriptsize[0.0,1.6]} & \makecell{1.7\\\scriptsize[0.2,3.1]} & \makecell{1.3\\\scriptsize[0.0,2.6]} & \makecell{0.3\\\scriptsize[0.0,1.0]} & \makecell{\textbf{2.3}\\\scriptsize[0.6,4.0]} \\
modify\_object & 303 & \makecell{5.9\\\scriptsize[3.3,8.6]} & \makecell{9.9\\\scriptsize[6.5,13.3]} & \makecell{14.2\\\scriptsize[10.3,18.1]} & \makecell{8.9\\\scriptsize[5.7,12.1]} & \makecell{23.8\\\scriptsize[19.0,28.6]} & \makecell{18.8\\\scriptsize[14.4,23.2]} & \makecell{28.4\\\scriptsize[23.3,33.5]} & \makecell{\textbf{29.7}\\\scriptsize[24.6,34.8]} & \makecell{7.3\\\scriptsize[4.3,10.2]} & \makecell{5.0\\\scriptsize[2.5,7.4]} & \makecell{8.3\\\scriptsize[5.2,11.3]} & \makecell{16.5\\\scriptsize[12.3,20.7]} & \makecell{19.5\\\scriptsize[15.0,23.9]} & \makecell{18.2\\\scriptsize[13.8,22.5]} \\
swap\_color & 312 & \makecell{2.2\\\scriptsize[0.6,3.9]} & \makecell{1.9\\\scriptsize[0.4,3.4]} & \makecell{1.6\\\scriptsize[0.2,3.0]} & \makecell{3.2\\\scriptsize[1.3,5.2]} & \makecell{3.2\\\scriptsize[1.3,5.2]} & \makecell{5.1\\\scriptsize[2.7,7.6]} & \makecell{2.6\\\scriptsize[0.8,4.3]} & \makecell{3.5\\\scriptsize[1.5,5.6]} & \makecell{3.2\\\scriptsize[1.3,5.2]} & \makecell{2.3\\\scriptsize[0.6,3.9]} & \makecell{\textbf{18.6}\\\scriptsize[14.3,22.9]} & \makecell{3.5\\\scriptsize[1.5,5.6]} & \makecell{1.6\\\scriptsize[0.2,3.0]} & \makecell{0.3\\\scriptsize[0.0,0.9]} \\
swap\_material & 303 & \makecell{0.0\\\scriptsize[0.0,0.0]} & \makecell{1.7\\\scriptsize[0.2,3.1]} & \makecell{2.0\\\scriptsize[0.4,3.5]} & \makecell{3.0\\\scriptsize[1.1,4.9]} & \makecell{2.0\\\scriptsize[0.4,3.5]} & \makecell{\textbf{4.3}\\\scriptsize[2.0,6.6]} & \makecell{1.3\\\scriptsize[0.0,2.6]} & \makecell{2.6\\\scriptsize[0.8,4.4]} & \makecell{0.3\\\scriptsize[0.0,1.0]} & \makecell{0.0\\\scriptsize[0.0,0.0]} & \makecell{3.0\\\scriptsize[1.1,4.9]} & \makecell{2.0\\\scriptsize[0.4,3.5]} & \makecell{1.0\\\scriptsize[0.0,2.1]} & \makecell{1.0\\\scriptsize[0.0,2.1]} \\
\midrule
\multicolumn{16}{l}{\emph{Quantity}} \\
modify\_count & 300 & \makecell{1.7\\\scriptsize[0.2,3.1]} & \makecell{0.7\\\scriptsize[0.0,1.6]} & \makecell{2.3\\\scriptsize[0.6,4.0]} & \makecell{2.0\\\scriptsize[0.4,3.6]} & \makecell{4.0\\\scriptsize[1.8,6.2]} & \makecell{3.7\\\scriptsize[1.5,5.8]} & \makecell{5.0\\\scriptsize[2.5,7.5]} & \makecell{\textbf{5.3}\\\scriptsize[2.8,7.9]} & \makecell{1.7\\\scriptsize[0.2,3.1]} & \makecell{0.7\\\scriptsize[0.0,1.6]} & \makecell{3.3\\\scriptsize[1.3,5.4]} & \makecell{4.7\\\scriptsize[2.3,7.1]} & \makecell{4.0\\\scriptsize[1.8,6.2]} & \makecell{3.3\\\scriptsize[1.3,5.4]} \\
modify\_presence & 312 & \makecell{0.6\\\scriptsize[0.0,1.5]} & \makecell{1.0\\\scriptsize[0.0,2.0]} & \makecell{1.6\\\scriptsize[0.2,3.0]} & \makecell{1.0\\\scriptsize[0.0,2.0]} & \makecell{1.0\\\scriptsize[0.0,2.0]} & \makecell{1.9\\\scriptsize[0.4,3.4]} & \makecell{\textbf{2.9}\\\scriptsize[1.0,4.7]} & \makecell{1.0\\\scriptsize[0.0,2.0]} & \makecell{1.3\\\scriptsize[0.0,2.5]} & \makecell{0.0\\\scriptsize[0.0,0.0]} & \makecell{0.6\\\scriptsize[0.0,1.5]} & \makecell{1.0\\\scriptsize[0.0,2.0]} & \makecell{0.6\\\scriptsize[0.0,1.5]} & \makecell{0.6\\\scriptsize[0.0,1.5]} \\
\midrule
\multicolumn{16}{l}{\emph{Spatial}} \\
modify\_spatial & 297 & \makecell{0.3\\\scriptsize[0.0,1.0]} & \makecell{\textbf{0.7}\\\scriptsize[0.0,1.6]} & \makecell{0.0\\\scriptsize[0.0,0.0]} & \makecell{0.3\\\scriptsize[0.0,1.0]} & \makecell{0.0\\\scriptsize[0.0,0.0]} & \makecell{0.0\\\scriptsize[0.0,0.0]} & \makecell{0.3\\\scriptsize[0.0,1.0]} & \makecell{0.3\\\scriptsize[0.0,1.0]} & \makecell{0.3\\\scriptsize[0.0,1.0]} & \makecell{\textbf{0.7}\\\scriptsize[0.0,1.6]} & \makecell{0.0\\\scriptsize[0.0,0.0]} & \makecell{0.0\\\scriptsize[0.0,0.0]} & \makecell{\textbf{0.7}\\\scriptsize[0.0,1.6]} & \makecell{0.0\\\scriptsize[0.0,0.0]} \\
swap\_spatial & 300 & \makecell{0.0\\\scriptsize[0.0,0.0]} & \makecell{0.0\\\scriptsize[0.0,0.0]} & \makecell{0.0\\\scriptsize[0.0,0.0]} & \makecell{0.0\\\scriptsize[0.0,0.0]} & \makecell{0.7\\\scriptsize[0.0,1.6]} & \makecell{0.7\\\scriptsize[0.0,1.6]} & \makecell{0.7\\\scriptsize[0.0,1.6]} & \makecell{\textbf{1.0}\\\scriptsize[0.0,2.1]} & \makecell{0.3\\\scriptsize[0.0,1.0]} & \makecell{0.0\\\scriptsize[0.0,0.0]} & \makecell{0.3\\\scriptsize[0.0,1.0]} & \makecell{0.0\\\scriptsize[0.0,0.0]} & \makecell{\textbf{1.0}\\\scriptsize[0.0,2.1]} & \makecell{0.3\\\scriptsize[0.0,1.0]} \\
\midrule
\multicolumn{16}{l}{\emph{Activity}} \\
modify\_action & 303 & \makecell{7.9\\\scriptsize[4.9,11.0]} & \makecell{17.2\\\scriptsize[12.9,21.4]} & \makecell{26.1\\\scriptsize[21.1,31.0]} & \makecell{22.1\\\scriptsize[17.4,26.8]} & \makecell{46.5\\\scriptsize[40.9,52.2]} & \makecell{21.1\\\scriptsize[16.5,25.7]} & \makecell{46.5\\\scriptsize[40.9,52.2]} & \makecell{\textbf{48.8}\\\scriptsize[43.2,54.5]} & \makecell{13.2\\\scriptsize[9.4,17.0]} & \makecell{6.9\\\scriptsize[4.1,9.8]} & \makecell{17.5\\\scriptsize[13.2,21.8]} & \makecell{33.3\\\scriptsize[28.0,38.6]} & \makecell{25.1\\\scriptsize[20.2,30.0]} & \makecell{31.4\\\scriptsize[26.1,36.6]} \\
modify\_pose & 312 & \makecell{3.8\\\scriptsize[1.7,6.0]} & \makecell{5.4\\\scriptsize[2.9,8.0]} & \makecell{4.5\\\scriptsize[2.2,6.8]} & \makecell{10.6\\\scriptsize[7.2,14.0]} & \makecell{9.6\\\scriptsize[6.3,12.9]} & \makecell{10.3\\\scriptsize[6.9,13.6]} & \makecell{10.3\\\scriptsize[6.9,13.6]} & \makecell{13.5\\\scriptsize[9.7,17.2]} & \makecell{7.1\\\scriptsize[4.2,9.9]} & \makecell{5.1\\\scriptsize[2.7,7.6]} & \makecell{\textbf{14.1}\\\scriptsize[10.2,18.0]} & \makecell{12.8\\\scriptsize[9.1,16.5]} & \makecell{5.1\\\scriptsize[2.7,7.6]} & \makecell{4.5\\\scriptsize[2.2,6.8]} \\
modify\_state & 438 & \makecell{0.5\\\scriptsize[0.0,1.1]} & \makecell{0.7\\\scriptsize[0.0,1.5]} & \makecell{1.1\\\scriptsize[0.1,2.1]} & \makecell{0.9\\\scriptsize[0.0,1.8]} & \makecell{2.1\\\scriptsize[0.7,3.4]} & \makecell{2.3\\\scriptsize[0.9,3.7]} & \makecell{\textbf{3.2}\\\scriptsize[1.5,4.8]} & \makecell{2.7\\\scriptsize[1.2,4.3]} & \makecell{1.1\\\scriptsize[0.1,2.1]} & \makecell{0.2\\\scriptsize[0.0,0.7]} & \makecell{0.5\\\scriptsize[0.0,1.1]} & \makecell{1.8\\\scriptsize[0.6,3.1]} & \makecell{0.7\\\scriptsize[0.0,1.5]} & \makecell{0.7\\\scriptsize[0.0,1.5]} \\
swap\_role & 135 & \makecell{0.0\\\scriptsize[0.0,0.0]} & \makecell{0.0\\\scriptsize[0.0,0.0]} & \makecell{0.0\\\scriptsize[0.0,0.0]} & \makecell{1.5\\\scriptsize[0.0,3.5]} & \makecell{1.5\\\scriptsize[0.0,3.5]} & \makecell{0.7\\\scriptsize[0.0,2.2]} & \makecell{1.5\\\scriptsize[0.0,3.5]} & \makecell{\textbf{3.0}\\\scriptsize[0.1,5.8]} & \makecell{2.2\\\scriptsize[0.0,4.7]} & \makecell{0.7\\\scriptsize[0.0,2.2]} & \makecell{2.2\\\scriptsize[0.0,4.7]} & \makecell{1.5\\\scriptsize[0.0,3.5]} & \makecell{\textbf{3.0}\\\scriptsize[0.1,5.8]} & \makecell{\textbf{3.0}\\\scriptsize[0.1,5.8]} \\
\bottomrule
\end{tabular}%
}
\end{table}

\end{landscape}
\begin{table}[H]
\centering
\caption{Per-edit-type Group score (\%) by caption verbosity, averaged across the 14 dual encoders of Table~\ref{tab:main-results}. Reveals which edit types degrade most (or least) as captions grow longer. Per-cell $n$ is per-tier instance count (one third of the column in Table~\ref{tab:app-per-edittype}).}
\label{tab:app-per-edittype-per-tier}
\small
\begin{tabular}{l r r r r}
\toprule
Edit type & $n$/tier & Short & Medium & Long \\
\midrule
\multicolumn{5}{l}{\emph{Identity}} \\
modify\_color & 206 & 35.3 & 34.7 & 30.5 \\
modify\_pattern & 101 & 16.1 & 15.2 & 14.4 \\
modify\_material & 161 & 10.2 & 10.0 & 8.3 \\
modify\_shape & 103 & 11.5 & 10.6 & 7.3 \\
modify\_transparency & 100 & 1.6 & 1.3 & 1.1 \\
modify\_object & 101 & 15.3 & 15.7 & 14.9 \\
swap\_color & 104 & 3.1 & 3.4 & 4.8 \\
swap\_material & 101 & 1.2 & 1.8 & 2.1 \\
\midrule
\multicolumn{5}{l}{\emph{Quantity}} \\
modify\_count & 100 & 3.4 & 3.1 & 2.5 \\
modify\_presence & 104 & 0.3 & 0.5 & 2.5 \\
\midrule
\multicolumn{5}{l}{\emph{Spatial}} \\
modify\_spatial & 99 & 0.4 & 0.1 & 0.3 \\
swap\_spatial & 100 & 0.4 & 0.4 & 0.3 \\
\midrule
\multicolumn{5}{l}{\emph{Activity}} \\
modify\_action & 101 & 27.2 & 27.3 & 23.4 \\
modify\_pose & 104 & 10.4 & 9.3 & 5.3 \\
modify\_state & 146 & 1.4 & 1.6 & 1.0 \\
swap\_role & 45 & 1.4 & 0.8 & 2.2 \\
\bottomrule
\end{tabular}
\end{table}

\begin{figure}[H]
\centering
\includegraphics[width=0.85\textwidth]{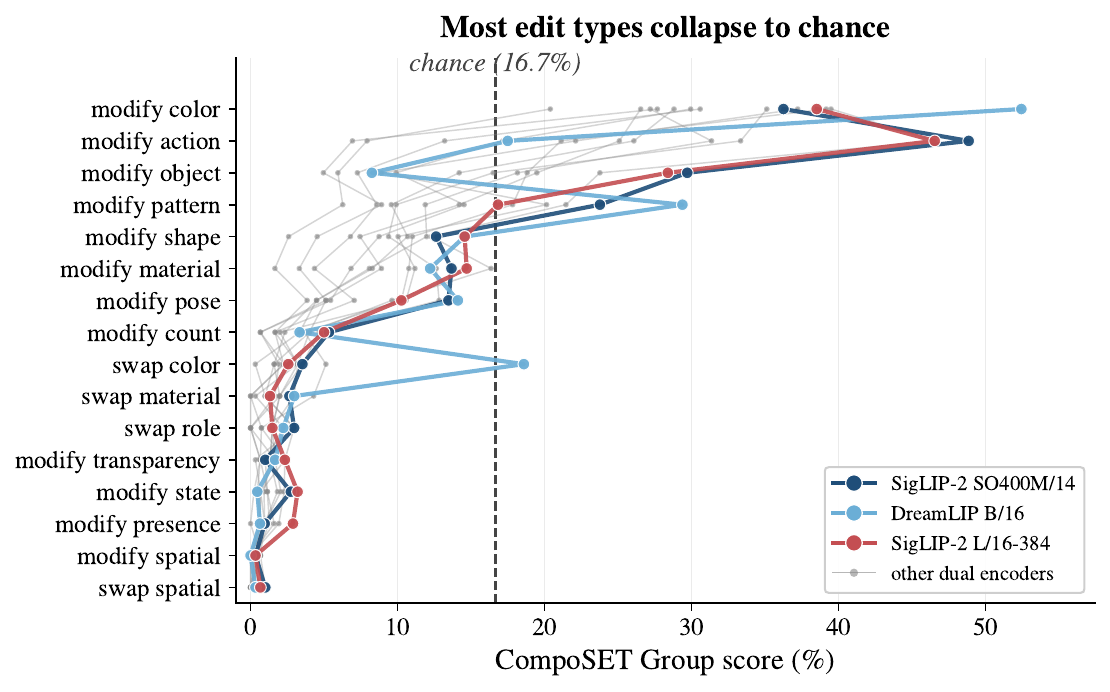}
\caption{\textbf{Most edit types are at or below chance.} CompoSET Group (\%) per edit type for the 14 dual encoders of Table~\ref{tab:main-results}, sorted by median across models (easiest at top). Top three by overall median: SigLIP-2 SO400M/14 (dark blue), DreamLIP B/16 (light blue), SigLIP-2 L/16-384 (red). Other 11 models in gray. Dashed: 16.67\% chance.}
\label{fig:per-edittype-range}
\end{figure}

\subsection{Cosine-Margin Analysis}
\label{app:cosine-margin}

For each (model, benchmark, pair), we compute the I2T \emph{margin} as $\text{sim}(\text{img}_a, \text{cap}_a) - \text{sim}(\text{img}_a, \text{cap}_b)$ and the analogous T2I margin. Margins are signed; positive on a correctly-decided pair, negative on an incorrectly-decided pair. Distributions are pooled across the 22 dual encoders.

\begin{figure}[H]
\centering
\includegraphics[width=\textwidth]{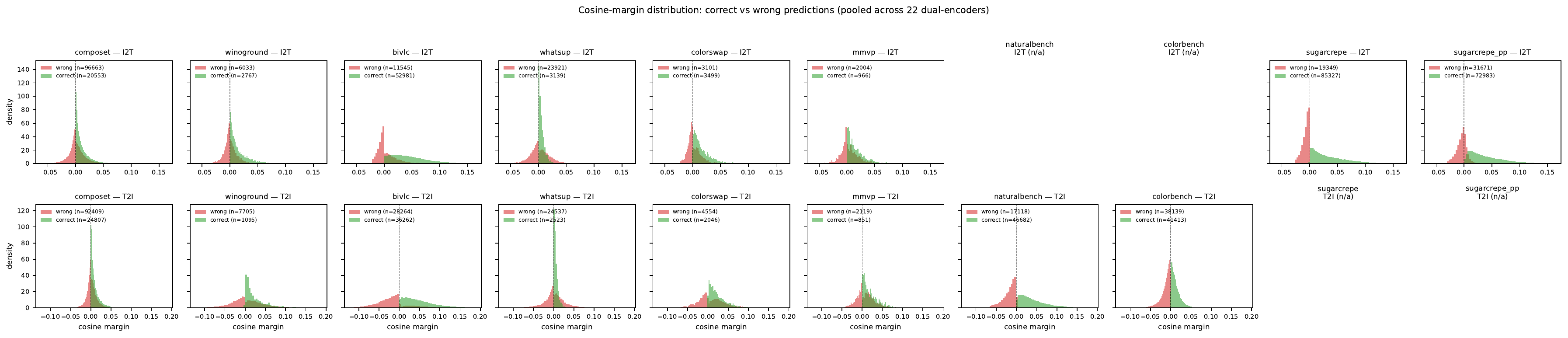}
\caption{Distribution of cosine margins on correct vs.\ wrong CompoSET decisions, pooled across the 22 dual encoders. The two distributions overlap heavily: even when the model decides correctly, the margin is small.}
\label{fig:cosine-margin-main}
\end{figure}

CompoSET correct-vs-wrong margins are substantially smaller than on prior benchmarks (Figure~\ref{fig:cosine-margin-main}). The best-margin dual-encoder on CompoSET I2T is DreamLIP B/16 (mean correct margin 0.0175 vs.\ wrong margin 0.0018, gap 0.0156); the worst is CLIP B/32 (gap 0.0044). For comparison, the same models on BiVLC achieve correct margins around 0.04--0.06. CompoSET pushes the dual-encoder closer to its decision boundary across the board.

\textbf{Per-scene scene-hard vs.\ edit-hard.} For each model and scene, we compute the mean and standard deviation of the I2T margin across the scene's $\sim$22 foils. Within each model, we label a scene \emph{scene-hard} if its mean margin falls in the bottom quartile (uniformly low margin: the scene itself is the obstacle) and \emph{edit-hard} if its margin standard deviation falls in the top quartile (some edits easy, others hard: the edit is the obstacle). The quartile thresholds fix each marginal rate at 25\% by construction; the informative quantities are the overlap between the labels and their structure across models and edit types, which we report below.

\emph{The two failure modes are nearly independent.} Across the 22 encoders, 4.7\% of model-scene pairs are both scene-hard and edit-hard, against 6.25\% expected if the labels were assigned independently. Scene-hardness and edit-hardness are distinct failure axes, not two views of one difficulty score.

\emph{Scene-hardness is consistent across models, with a model-specific component.} The labels are defined per model, so they describe a joint property of data and model rather than an intrinsic dataset annotation. The rankings they induce, however, agree well beyond chance: per-scene mean-margin rankings correlate at mean pairwise Spearman $\rho=0.56$ (range 0.26--0.84); scene-hard sets overlap at mean Jaccard 0.36, against 0.14 for independent quartile draws; and 17 of 80 scenes are scene-hard for at least 11 of the 22 encoders, while 19 are scene-hard for none. A two-way decomposition attributes 53\% of per-scene mean-margin variance to scene identity and 10\% to model identity, leaving 37\% to their interaction. Encoders that share a training recipe agree slightly more than those that do not (same-family $\rho=0.62$ vs.\ cross-family $0.55$).

\emph{Edit-hardness does not reduce to edit-type composition.} A scene could rank as edit-hard merely because it contains edit types that are hard everywhere. It does not: edit type alone explains 3.7\% of margin variance, while the scene$\times$edit-type interaction explains a further 16\% --- the same edit type is easy in one scene and hard in another. Residualizing out per-edit-type means leaves the per-model edit-hard ranking essentially unchanged (mean Spearman $\rho=0.95$; top-quartile overlap 89\%), and scenes containing the least pixel-local edit types (\texttt{swap\_role}, \texttt{modify\_pose}, \texttt{modify\_action}) are not measurably more edit-hard than scenes without them (Mann-Whitney $p=0.33$).

\emph{Foil identity carries most of the signal.} 94.7\% of per-foil success-rate variance lies within scenes; 5.3\% lies between. A one-foil-per-base benchmark samples only the between-scene component. The many-foils-per-base structure is what makes the decomposition above measurable at all.

\subsection{Generative-VLM Slice}
\label{app:gen-vlm}

We separately evaluate generative VLMs on a stratified 200-instance CompoSET subset; results are not directly comparable to the dual-encoder pool because the scoring primitive differs.

\paragraph{Open-weight (caption log-likelihood).} BLIP-2 (OPT-2.7B)~\cite{li2023blip2}, LLaVA-1.6 (Mistral-7B)~\cite{liu2024llavanext}, and Qwen2-VL (7B Instruct)~\cite{wang2024qwen2vl}. Each candidate caption is scored by its mean per-token log-likelihood under the image (length-normalized; without normalization, BLIP-2 on SugarCrepe inflates I2T by $\sim$11~pp). I2T is the primary metric for this slice~\cite{thrush2022winoground,yuksekgonul2023bow,hsieh2023sugarcrepe}: across-image log-likelihoods carry image-side variance unrelated to caption-image fit. Qwen2-VL on CompoSET shows the asymmetry: 36.5\% I2T vs.\ 0.5\% T2I.

\paragraph{Frontier API (prompted pairwise selection).} Gemini 2.5 Flash~\cite{gemini25flash}, Claude Haiku 4.5~\cite{claudehaiku45}, and Claude Sonnet 4.6. Each $2\times2$ instance issues four API calls (two I2T, two T2I) asking the model to choose between two candidate options. Frontier models show a position bias under this prompting~\cite{wang2024llmjudges}: they pick whichever option appeared in a fixed position, partly independent of correctness. We counterbalance every call by issuing it twice with the option order swapped and averaging the binary correctness. Without counterbalancing, Gemini Flash on CompoSET T2I scored 76.5\% when the correct image was first but only 30.0\% when it was second; Haiku 4.5 showed a symmetric bias toward the second option. Cost: about \$10 total on the 200-instance subset.

Table~\ref{tab:gen-vlm-composet} reports CompoSET I2T, T2I, and Group on the 200-instance subset for both slices. Claude Sonnet 4.6 reaches 42.5\% Group, about three times the dual-encoder ceiling; Gemini 2.5 Flash and Claude Haiku 4.5 reach 12.0\% and 13.0\%. Open-weight models collapse on T2I (Qwen2-VL 0.5\%, LLaVA-1.6 1.5\%) because across-image log-likelihoods carry image-side variance unrelated to caption-image fit; their I2T column shows the same models score 30--37\% when ranking captions for a fixed image. Per-verbosity Group (Table~\ref{tab:gen-vlm-tier}) shows a consistent medium peak across the three frontier models: 44.4 / 46.5 / 36.4 (Sonnet), 15.9 / 19.7 / 13.6 (Gemini), 11.1 / 19.7 / 7.6 (Haiku). With $\sim$67 instances per tier the per-tier intervals are wide, but the cross-model pattern holds.

\begin{table}[H]
\centering
\caption{\textbf{Generative VLMs on CompoSET (200-instance subset).} Open-weight rows scored by length-normalized caption log-likelihood. Frontier rows scored by prompted pairwise selection with option-order counterbalancing. Best Group in \textbf{bold}.}
\label{tab:gen-vlm-composet}
\small
\begin{tabular}{l r r r}
\toprule
Model & I2T & T2I & Group \\
\midrule
\multicolumn{4}{l}{\emph{Open-weight (caption log-likelihood)}} \\
BLIP-2 (OPT-2.7B)               & 5.0  & 11.0 & 2.5 \\
LLaVA-1.6 (Mistral-7B)          & 30.5 & 3.0  & 1.5 \\
Qwen2-VL (7B Instruct)          & 36.5 & 0.5  & 0.5 \\
\midrule
\multicolumn{4}{l}{\emph{Frontier API (prompted pairwise, counterbalanced)}} \\
Gemini 2.5 Flash                & 64.5 & 16.0 & 12.0 \\
Claude Haiku 4.5                & 46.5 & 17.0 & 13.0 \\
Claude Sonnet 4.6               & 71.5 & 47.0 & \textbf{42.5} \\
\bottomrule
\end{tabular}
\end{table}

\begin{table}[H]
\centering
\caption{\textbf{Frontier VLMs peak at medium verbosity on CompoSET.} Group score (\%) per caption verbosity tier, $\sim$67 instances per tier.}
\label{tab:gen-vlm-tier}
\small
\begin{tabular}{l r r r}
\toprule
Model & Short & Medium & Long \\
\midrule
Gemini 2.5 Flash    & 15.9 & 19.7 & 13.6 \\
Claude Haiku 4.5    & 11.1 & 19.7 & 7.6  \\
Claude Sonnet 4.6   & 44.4 & 46.5 & 36.4 \\
\bottomrule
\end{tabular}
\end{table}

\end{document}